\documentclass[sn-mathphys-num]{sn-jnl}

\usepackage{lmodern}
\usepackage{ulem}

\usepackage{amsmath}
\usepackage{mathtools}
\usepackage{amssymb}
\usepackage{mathrsfs}
\usepackage{array} 

\usepackage{siunitx}
\DeclareSIUnit\nounit{\relax}

\usepackage{pgfplots}
\pgfplotsset{compat=1.18}
\usepackage{graphics}
\usepackage{graphicx}
\usepackage{tikz}
\usetikzlibrary{shapes.geometric, arrows}

\usepackage{hyperref}
\usepackage[nameinlink, capitalize]{cleveref}
\crefname{equation}{}{}

\usepackage[nolist]{acronym}

\usepackage[labelfont=bf, figurename=Fig.]{caption}
\usepackage{subcaption}
\newcolumntype{C}[1]{>{\centering\arraybackslash\hspace{0pt}}p{#1}}
\usepackage{multirow}
\usepackage{framed}
\usepackage{booktabs}
\usepackage{enumitem}

\usepackage{color}
\usepackage{tcolorbox}

\newcommand{\bm}[1]{\boldsymbol{#1}}

\newcommand{\indexOfNum}[1]{{n_{\textrm{#1}}}}
\newcommand{\minIndex}[1]{{#1_{\textrm{min}}}}
\newcommand{\maxIndex}[1]{{#1_{\textrm{max}}}}
\newcommand{\meanSup}[1]{{#1^{\textrm{mean}}}}

\newcommand{\ffnn}{f_{\textrm{N}}}
\newcommand{\ffnnInput}{\ffnn^{\textrm{in}}}
\newcommand{\ffnnOutput}{\ffnn^{\textrm{out}}}
\newcommand{\numLayers}{\indexOfNum{\textrm{L}}}
\newcommand{\numNeurons}{\indexOfNum{\textrm{n}}}

\newcommand{\ansatzU}{\mathcal{U}}
\newcommand{\intHBCAnsatzU}{\tilde{\ansatzU}_{\textrm{hbc}}}
\newcommand{\intNormAnsatzU}{\tilde{\ansatzU}_{\textrm{n}}}

\newcommand{\lossF}{F^{\textrm{L}}}
\newcommand{\lossFC}{\lossF_{\textrm{C}}}
\newcommand{\lossFN}{\lossF_{\textrm{N}}}
\newcommand{\lossFd}{\lossF_{\textrm{d}}}

\newcommand{\lambdaC}{\lambda_{\textrm{C}}}
\newcommand{\lambdaN}{\lambda_{\textrm{N}}}
\newcommand{\lambdad}{\lambda_{\textrm{d}}}

\newcommand{\dataSet}{\vec{T}}
\newcommand{\dataSetC}{\vec{T}_{\textrm{C}}}
\newcommand{\dataSetN}{\vec{T}_{\textrm{N}}}
\newcommand{\dataSetd}{\vec{T}_{\textrm{d}}}

\newcommand{\PINNSup}[1]{{#1^{\textrm{PINN}}}}

\newcommand{\MAE}{\text{MAE}}
\newcommand{\ARE}{\text{ARE}}
\newcommand{\MARE}{\text{MARE}}
\newcommand{\LTwo}{\text{L}^{2}}
\newcommand{\rLTwo}{\text{rL}^{2}}
\newcommand{\RD}{\text{RD}}

\newcommand{\numTests}{\indexOfNum{tests}}
\newcommand{\trueSup}[1]{{#1^{\textrm{true}}}}
\newcommand{\identifiedSup}[1]{{#1^{\textrm{identified}}}}
\newcommand{\likelihood}{L_{\textrm{d}}}

\newcommand{\numNodes}{\indexOfNum{nodes}}
\newcommand{\FEMSup}[1]{{#1^{\textrm{FEM}}}}

\newcommand{\bodyR}{\mathcal{B}_{\textrm{R}}}
\newcommand{\motionR}{\bm{\chi}_\textrm{R}}
\newcommand{\rohR}{\rho_{\textrm{R}}}
\newcommand{\normalR}{\ten{n}_{\textrm{R}}}
\newcommand{\gammaR}{\Gamma_{\textrm{R}}}
\newcommand{\gammaDirichletR}{\gammaR^{\textrm{D}}}
\newcommand{\gammaNeumannR}{\gammaR^{\textrm{N}}}
\newcommand{\psiR}{\psi_{\textrm{R}}}
\newcommand{\psiVolR}{\psiR^{\textrm{vol}}}
\newcommand{\psiIsoR}{\psiR^{\textrm{iso}}}
\newcommand{\deviatoricIndex}[1]{{#1_{\textrm{D}}}}
\newcommand{\numKappa}{n_{\kappa}}

\newcommand{\numC}{\indexOfNum{\textrm{C}}}
\newcommand{\numD}{\indexOfNum{\textrm{D}}}
\newcommand{\numN}{\indexOfNum{\textrm{N}}}
\newcommand{\numd}{\indexOfNum{\textrm{d}}}
\newcommand{\numu}{\indexOfNum{\textrm{u}}}

\newcommand{\vecF}{\vec{F}}
\newcommand{\vecFC}{\vecF_{\textrm{C}}}
\newcommand{\vecFD}{\vecF_{\textrm{D}}}
\newcommand{\vecFN}{\vecF_{\textrm{N}}}

\newcommand{\uState}{\vec{u}^{\textrm{s}}}
\newcommand{\hatUState}{\hatvec{u}^{\textrm{s}}}
\newcommand{\uModelState}{\vec{u}^{\textrm{m}}}
\newcommand{\uModelStateC}{\uModelState_{\textrm{C}}}
\newcommand{\uModelStateD}{\uModelState_{\textrm{D}}}
\newcommand{\uModelStateN}{\uModelState_{\textrm{N}}}

\DeclareMathOperator*{\argmin}{arg\,min}
\DeclareMathOperator*{\argmax}{arg\,max}

\newcommand{\norm}[1]{\left\lVert#1\right\rVert}  

\newcommand{\sca}[1]{ \ensuremath{ \mathrm{#1} } }
\newcommand{\barsca}[1]{ \ensuremath{ \bar{ \sca{#1} } } }
\newcommand{\hatsca}[1]{ \ensuremath{ \widehat{ \sca{#1} } } }

\renewcommand{\vec}[1]{ \ensuremath{ \mathbf{#1} } }
\newcommand{\barvec}[1]{\ensuremath{ \bar{ \vec{#1} } } }
\newcommand{\hatvec}[1]{\ensuremath{ \widehat{ \vec{#1} } } }

\newcommand{\ten}[1]{\ensuremath{ \boldsymbol{ \mathsf{#1} } } }
\newcommand{\barten}[1]{\ensuremath{ \bar{ \ten{#1} } } }

\newcommand{\elm}[1]{{\, \in \mathbb{R}}^{#1}}

\newcommand{\elmm}[2]{{\, \in \mathbb{R}}^{#1 \times #2}}

\newcommand{\half}{\frac{1}{2}}

\renewcommand{\d}[1]{\text{$\hspace{0.1cm}$d $\hspace{-0.11cm}#1$}}

\newcommand{\tr}[1]{\text{tr$\left(#1\right)$}}

\renewcommand{\det}[1]{\text{$\hspace{0.1cm}$det$\left(#1\right)$}}

\renewcommand{\it}{\ensuremath{ ^{-T} }}

\begin{acronym}
\acro{SHM}[SHM]{structural health monitoring}
\acro{DIC}[DIC]{digital image correlation}
\acro{ESPI}[ESPI]{electronic speckle pattern interferometry}
\acro{PDE}[PDE]{partial differential equation}
\acro{MCMC}[MCMC]{Markov chain Monte Carlo}
\acro{PDF}[PDF]{probability density function}
\acro{NLS}[NLS]{nonlinear least-squares}
\acro{ANN}[ANN]{artificial neural network}
\acro{FFNN}[FFNN]{feed-forward neural network}
\acro{PINN}[PINN]{physics-informed neural network}
\acrodefplural{PINNs}[PINNs]{Physics-informed neural networks}
\acro{FE}[FE]{finite element}
\acro{FEM}[FEM]{finite element method}
\acro{BC}[BC]{boundary condition}
\acro{NLS-FEM}[NLS-FEM]{nonlinear least-squares finite element method}
\acro{VFM}[VFM]{virtual fields method}
\acro{RE}[RE]{relative error}
\acro{MAE}[MAE]{mean absolute error}
\acro{rL2}[$\text{rL}^{2}$]{relative $\text{L}^{2}$ norm}
\acro{RD}[RD]{relative deviation}
\acro{ARE}[ARE]{absolute relative error}
\acro{MARE}[MARE]{mean absolute relative error}
\acro{SEM}[SEM]{standard error of the mean}
\acrodefplural{SEMs}[SEMs]{standard errors of the means}
\acro{GPU}[GPU]{graphics processing unit}
\acro{CPU}[CPU]{central processing unit}
\end{acronym}

\begin{document}

\title{Deterministic and statistical calibration of constitutive models from full-field data with parametric physics-informed neural networks}

\author*[1]{\fnm{David} \sur{Anton}}\email{d.anton@tu-braunschweig.de}
\author[2]{\fnm{Jendrik-Alexander} \sur{Tröger}}\email{jendrik-alexander.troeger@tu-clausthal.de}
\author[1]{\fnm{Henning} \sur{Wessels}}\email{h.wessels@tu-braunschweig.de}
\author[3]{\fnm{Ulrich} \sur{Römer}}\email{u.roemer@tu-braunschweig.de}
\author[4]{\fnm{Alexander} \sur{Henkes}}\email{ahenke@ethz.ch}
\author[2]{\fnm{Stefan} \sur{Hartmann}}\email{stefan.hartmann@tu-clausthal.de}


\affil[1]{\orgdiv{Institute of Applied Mechanics}, \orgname{Technische Universität Braunschweig}, \orgaddress{\street{Pockelsstraße 3}, \city{Braunschweig}, \postcode{38106}, \country{Germany}}}

\affil[2]{\orgdiv{Institute of Applied Mechanics}, \orgname{Clausthal University of Technology}, \orgaddress{\street{Adolph-Roemer-Straße 2A}, \city{Clausthal-Zellerfeld}, \postcode{38678}, \country{Germany}}}

\affil[3]{\orgdiv{Institute for Acoustics and Dynamics}, \orgname{Technische Universität Braunschweig}, \orgaddress{\street{Langer Kamp 19}, \city{Braunschweig}, \postcode{38106}, \country{Germany}}}

\affil[4]{\orgdiv{Computational Mechanics Group}, \orgname{Eidgenössische Technische Hochschule Zürich}, \orgaddress{\street{Tannenstrasse 3}, \city{Zürich}, \postcode{8092}, \country{Switzerland}}}

\abstract{
The calibration of constitutive models from full-field data has recently gained increasing interest due to improvements in full-field measurement capabilities. In addition to the experimental characterization of novel materials, continuous structural health monitoring is another application that is of great interest. However, monitoring is usually associated with severe time constraints, difficult to meet with standard numerical approaches. Therefore, parametric \acp{PINN} for constitutive model calibration from full-field displacement data are investigated. In an offline stage, a parametric \ac{PINN} can be trained to learn a parameterized solution of the underlying partial differential equation. In the subsequent online stage, the parametric \ac{PINN} then acts as a surrogate for the parameters-to-state map in calibration. We test the proposed approach for the deterministic least-squares calibration of a linear elastic as well as a hyperelastic constitutive model from noisy synthetic displacement data. We further carry out Markov chain Monte Carlo-based Bayesian inference to quantify the uncertainty. A proper statistical evaluation of the results underlines the high accuracy of the deterministic calibration and that the estimated uncertainty is valid. Finally, we consider experimental data and show that the results are in good agreement with a finite element method-based calibration. Due to the fast evaluation of \acp{PINN}, calibration can be performed in near real-time. This advantage is particularly evident in many-query applications such as Markov chain Monte Carlo-based Bayesian inference. 
}


\maketitle


\section{Introduction}\label{sec:introduction}

The calibration of constitutive models is a major research field in computational as well as experimental solid mechanics and has a wide range of applications in practice. The interest in appropriate methods for constitutive model calibration recently increased further with the improvement of full-field measurement capabilities and the associated increase in available full-field displacement data. Probably the most obvious application in the context of experimental solid mechanics is the characterization of novel materials from experimental data. Another application that is gaining increasing interest is continuous \acf{SHM} \cite{chang_reportFirstSHMWorkshop_1998, entezami_structuralHealthMonitoring_2021}. Material parameters directly reflect the resistance to external impacts and indicate damage and material degradation and thus provide crucial information for the assessment of existing structures. Since in \ac{SHM} stress data is typically not accessible, the material parameters of interest must be identified from displacement or strain data, measured, e.g., by \acf{DIC} \cite{sutton_imageCorrelation_2009} or \acf{ESPI} \cite{yang_strainMeasurementBy3DESPI_2003}, respectively. 

The connection between constitutive model parameters and the measured full-field data is then established by the inverse solution of the parametric mechanical model. Traditionally, this inverse problem is solved by numerical methods, such as the \acf{NLS-FEM}, see, for instance, \cite{mahnken_parameterIdentificationFEM_1996,rose_parameterIdentificationUsingFullFieldMeasurements_2020}, or the \ac{VFM} \cite{avril_overviewIdentificationMethods_2008,martins_comparisonInverseIdentificationStrategies_2018}. While both \ac{NLS-FEM} and \ac{VFM} are well established in experimental mechanics, their application in \ac{SHM} is oftentimes prohibitive since their computational costs do not meet the severe time constraints in online applications. Thus, there is great interest in methods that are suitable for repeated calibration in the laboratory or in online applications.

Recently, it has been shown that \acfp{PINN} \cite{raissi_physicsInformedNeuralNetworks_2019} are particularly suited for solving inverse problems. \Acp{PINN} are a framework for solving forward and inverse problems involving nonlinear \acp{PDE} from the field of physics-informed machine learning \cite{karniadakis_physicsInformedMachineLearning_2021}. The idea behind this method goes back to the 1990s \cite{psichogios_neuralNetworkToProcessModeling_1992, lagaris_artificialNeuralNetworksForSolvingPDEs_1998}, but it became applicable only recently due to developments in automatic differentiation \cite{baydin_automaticDifferentiation_2018}, software frameworks, such as TensorFlow \cite{abadi_TensorFlow_2015} and PyTorch \cite{paszke_PyTorch_2019}, and more powerful hardware. The main advantages of \acp{PINN} are a straightforward inclusion of training data and their use as a continuous ansatz function. Thanks to the latter, all quantities can be computed directly on the sensor locations, bypassing the need for interpolation as, e.g., in \ac{FEM}-based calibration approaches.

In general, most numerical methods for calibrating constitutive models from full-field data can be classified into \textit{reduced} and \textit{all-at-once} approaches, see \cite{roemer_modelCalibrationInSolidMechanics_2024} for a recent review. Therein, an unifying framework for model calibration in computational solid mechanics has been developed. The reduced approach assumes that a parameters-to-state map exists, which is provided, e.g., by a \ac{PINN} or a \ac{FE} simulation. In contrast, in the all-at-once approach, the state and the model parameters are inferred simultaneously. For \acp{PINN} as well as other numerical methods, it is possible to formulate the calibration problem both in the reduced as well as in the all-at-once setting. 

In the literature, most contributions focusing on parameter identification with \acp{PINN} are associated with the all-at-once approach. Such formulations are also referred to as inverse \acp{PINN}. In \cite{shukla_PINNsForMicrostructuralProperties_2022,rojas_PINNsParameterIdentificationDamageModel_2023,zhang_PINNsAnalysesInternalStructures_2022,haghighat_PINNsInSolidMechanics_2021,hamel_PINNsCalibratingConstitutiveModels_2022,zhang_PINNsNonhomogeneousMaterialIdentification_2020}, inverse \acp{PINN} have been applied to parameter identification from full-field displacement data. However, many of the assumptions made therein do not match the conditions of real-world applications. This mainly concerns the magnitude and quality of the measured displacements. Some references, such as \cite{haghighat_PINNsInSolidMechanics_2021}, even consider the availability of full-field stress data for identification, which in practice must be considered as unknown. In earlier work, some of the authors have further developed inverse \acp{PINN} towards parameter identification in a realistic regime \cite{anton_PINNsMaterialModelCalibration_2023}, both concerning the magnitude of the material parameters as well as the noise level of the displacement data. Nevertheless, a severe restriction of inverse \acp{PINN} remains. In principle, they must be trained from scratch each time new measurements become available. This involves high computational costs and is a significant disadvantage when it comes to repeated online calibration or standardized material tests, where the setup basically remains the same.

In this contribution, we therefore focus on \acp{PINN} in a reduced approach. In an offline stage, the \ac{PINN} is trained to learn a parameterized solution of the underlying parametric \ac{PDE} within a predefined range of material parameters. For this purpose, the material parameters are considered as additional inputs to the \ac{PINN}, such that the predicted displacement no longer depends on the spatial position only, but also on the material parameters. To speed up the training process and to make it more robust, we suggest to include some data in the training process. This data may be generated by high-fidelity \ac{FE} simulations. In the subsequent online stage, the pre-trained \ac{PINN} then acts as a surrogate for the parameters-to-state map in calibration. This special variant of \acp{PINN}, known as \textit{parametric \acp{PINN}}, has already been deployed for thermal analysis of a laser powder bed fusion process \cite{hosseini_ParametricPINNsPowderBedFusion_2023}, magnetostatic problems \cite{beltranpulido_ParametricPINNsMagnetostaticProblems_2022}, or for the optimization of an airfoil geometry \cite{sun_ParametricPINNsAirfoilGeometry_2023}. To the best of our knowledge, parametric \acp{PINN} have not yet been used for the calibration of constitutive models in solid mechanics using real-world experimental data. Building up on our results reported in \cite{roemer_modelCalibrationInSolidMechanics_2024}, we statistically evaluate the accuracy of the parametric \acp{PINN} for the calibration of constitutive models from noisy synthetic full-field data, extend the study to hyperelastic materials and consider experimental data.

We demonstrate that the parametric \ac{PINN} approach enables an accurate and efficient model calibration and uncertainty quantification of the inferred material parameters in online applications, even though up to $\mathcal{O}(10^{4})$ forward model evaluations are required. To illustrate this, we first consider the constitutive models for both small strain linear elasticity and finite strain hyperelasticity and perform a re-identification of the material parameters from noisy synthetic displacement data. In the deterministic setting, a \ac{NLS} problem is solved. A statistical evaluation of the results shows that the point estimates obtained by solving the \ac{NLS} problem deviate only marginally from the true material parameters. We further treat the material parameters as random variables, conduct Bayesian statistical inference and quantify the uncertainty in the estimated material parameters. The posterior distribution of the material parameters is determined by carrying out a \ac{MCMC} analysis. In order to validate the quantified uncertainty from a frequentist point of view, we perform a coverage test. The results for the statistical calibration show that the estimated uncertainties are also valid. 
In addition to the synthetic data, we calibrate the constitutive model for small strain linear elasticity using experimental full-field displacement data obtained from a tensile test. We demonstrate that the calibration with a parametric \ac{PINN} shows good results compared to using \ac{FEM} for both the deterministic as well as the statistical setting.


In summary, the advantages of using parametric \acp{PINN} as surrogates of the parameters-to-state map in the context of constitutive model calibration are:
\begin{itemize}
    \item \textbf{Parametric \acp{PINN} allow for a near real-time calibration.} Once a \ac{PINN} has been trained in the offline stage, the evaluation of the parameters-to-state map in the online stage is very cheap. This is a clear advantage, especially when used in many-query approaches such as the deterministic \ac{NLS} approach or the statistical \ac{MCMC} analysis.
    \item \textbf{Parametric \acp{PINN} are continuous ansatz functions.} No interpolation between the sensor locations and the numerical discretization is required for calibration.
    \item \textbf{Data can be easily integrated to speed up training.} Data is not necessary for training, but can speed up the training process and make it more robust. As with projection-based reduced order modeling approaches \cite{agarwal_ROMForHydrogels_2024, stoter_DEIMDrivenROM_2022}, such data may arise from snapshots of a high-fidelity \ac{FE} model.
\end{itemize}
To support the advantages mentioned above and to increase the acceptance of parametric \acp{PINN} in the context of constitutive model calibration, the present study aims towards the following key contributions:
\begin{itemize}
    \item \textbf{We use parametric \acp{PINN} for uncertainty quantification.} The parametric \ac{PINN} is used as surrogate of the parameters-to-state map within a \ac{MCMC} analysis and provides us with the posterior probability density of the parameters of interest. 
    \item \textbf{We perform a statistical evaluation of the numerical results.} To validate the estimated uncertainty in the Bayesian statistical setting from a frequentist point of view, we perform a coverage test.
    \item \textbf{We consider real-world experimental displacement data.} We calibrate a linear elastic constitutive model using experimental data measured in a tensile test.
\end{itemize}
To the best of the authors knowledge, the above mentioned contributions in connection with parametric \acp{PINN} have not yet been considered in the literature.

\noindent The code for our numerical tests including the data generation, the training and validation of parametric \acp{PINN}  as well as the calibration methods is implemented in the Python programming language. The \ac{PINN} implementation is mainly based on the PyTorch framework \cite{paszke_PyTorch_2019}. The code for the \ac{FE} data generation is built on top of the FEniCSx project \cite{baratta_DOLFINx_2023}. Our research code is open source and available both on GitHub and Zenodo \cite{anton_codeParametricPINNsCalibration_2024}. In addition, we also published the experimental data set on Zenodo \cite{troeger_DICMeasurementLinearElasticSteel_2024}.

The remainder of this paper is structured as follows: In \cref{sec:mechanics}, the balance of linear momentum and the considered constitutive models are recapitulated. We then provide a brief introduction to \acp{ANN} and parametric \acp{PINN} in \cref{sec:PPINNs}. In this section, we also elaborate on normalization steps necessary for real-world applications. In \cref{sec:problem_statement}, the calibration problem both in the deterministic as well as the Bayesian statistical setting are formulated. Subsequently, in \cref{sec:results_synthetic} and \cref{sec:results_experimental}, we provide the results for our numerical tests including both synthetic and experimental full-field data, respectively. Finally, we conclude our investigations with a critical discussion and point out possible directions of future work in \cref{sec:conclusion}.


\section{Solid mechanics preliminaries}\label{sec:mechanics}

The relationship between the measured displacements of a body and the material parameters is defined by the framework of solid mechanics. In the following, we briefly recapitulate the balance of linear momentum and elaborate on the constitutive models for both small strain linear elasticity and finite strain hyperelasticity. For a more in-depth introduction to solid mechanics, the reader is referred to standard text books, e.g., \cite{holzapfel_nonlinearSolidMechanics_2000,wriggers_nonlinearFEM_2008}.

\subsection{Fundamental equations}\label{subsec:mechanics_fundamental_equations}

The displacement of a material point $\vec{X} \in \bodyR$ in the undeformed reference configuration $\bodyR$ (denoted by subscript $_{\textrm{R}}$) is defined by 
\begin{equation}\label{eq:displacement}
    \vec{u} (\vec{X},t) = \vec{x} - \vec{X} = \motionR(\vec{X},t) - \vec{X},
\end{equation}
where the vector $\vec{x} \in \mathcal{B}$ corresponds to the position of a material point in the deformed configuration $\mathcal{B}$ at time $t$ and $\motionR(\vec{X},t)$ represents the motion. 
In the following, the explicit time dependence is omitted for brevity. Furthermore, both the undeformed reference configuration $\bodyR$ and the deformed configuration $\mathcal{B}$ are modeled as a subset of the physical Euclidean space $\mathbb{E}^{3}$ with orthonormal basis vectors. Then, $\mathbb{E}^{3}$ can be identified with the common three-dimensional vector space $\mathbb{R}^{3}$. More information on the geometrical treatment of continuum mechanics can be found in \cite{marsden_mathematicalFoundationsElasticity_1983,lychev_differentialGeometryContinuumMechanics_2019}.

In the reference configuration $\bodyR$, the balance of linear momentum in its strong form and in static equilibrium states
\begin{equation}\label{eq:balance_momentum}
    \operatorname{Div} \ten{P}(\vec{X}; \bm{\kappa}) + \rohR(\vec{X}) \, \vec{b}(\vec{X}) = \vec{0}, \; \vec{X} \in \bodyR.
\end{equation}
Here, $\operatorname{Div}$ denotes the divergence operator with respect to the coordinates $\vec{X}$ and $\ten{P}$ represents the first Piola-Kirchhoff stress tensor. The density in the reference configuration is denoted by $\rohR$ and $\vec{b}$ are accelerations caused, for instance, by gravity. Equation \cref{eq:balance_momentum} needs to be satisfied for all points $\vec{X}$ inside the domain $\bodyR$. The stress depends on the displacement $\vec{u}$ via the strains and is parameterized by a set of material parameters $\bm{\kappa} \elm{n_\kappa}$. The semicolon indicates parameterization of $\ten{P}$ in $\bm{\kappa}$. 

The \ac{PDE} \cref{eq:balance_momentum} is complemented by a set of Dirichlet and Neumann boundary conditions
\begin{subequations}\label{eq:bcs}
\begin{align}
    \vec{u}(\vec{X}) &= \bar{\vec{u}}, \; \vec{X} \in \gammaDirichletR, \label{eq:bcs_dirichlet}\\
    \ten{P}(\vec{X}; \bm{\kappa}) \cdot \normalR(\vec{X}) &= \barvec{t}, \; \vec{X} \in \gammaNeumannR, \label{eq:bcs_neumann}
\end{align}
\end{subequations}
with $\gammaDirichletR$ and $\gammaNeumannR$ denoting the complementary surfaces of the boundary $\gammaR = \partial \bodyR$, with $\gammaDirichletR\, \cup\, \gammaNeumannR = \gammaR$. Furthermore, $\barvec{u}$ and $\barvec{t}$ are the prescribed displacements and tractions on the boundaries, respectively, and $\normalR$ is the normal vector on the outer surface of the reference configuration.

The system of equations arising from \cref{eq:balance_momentum,eq:bcs} is closed by the kinematics and a constitutive model describing the stress state as a function of strain, parameterized in the  material parameters $\bm{\kappa}$. In the following, we briefly recall the kinematics and constitutive equations for linear elasticity and hyperelasticity.

\subsection{Linear elasticity}\label{subsec:linear_elasticity}

For linear, isotropic elasticity and small strains, the constitutive model states
\begin{equation}\label{eq:model_linearelasticity}
    \ten{\sigma}(\ten{\epsilon}; \bm{\kappa}) =  K \, \tr{\ten{\epsilon}} \ten{I} + 2 G \deviatoricIndex{\ten{\epsilon}}, 
\end{equation}
where $\ten{\sigma}$ is the Cauchy stress tensor, $\ten{I}$ is the second-order identity tensor and $\text{tr}$ is the trace operator. Note that in the linear elastic theory, it is assumed that $\ten{P}\approx\ten{\sigma}$ in \cref{eq:balance_momentum,eq:bcs}. The linear strain tensor $\ten{\epsilon}$ is defined as
\begin{equation}\label{eq:kinematic_linear_elasticity}
    \ten{\epsilon} = 
    \frac{1}{2} 
    \Bigl( 
    \operatorname{Grad}{\vec{u}(\vec{X})} 
    + (\operatorname{Grad}{\vec{u}(\vec{X})})^{\top} 
    \Bigr),
\end{equation}
where the gradient $\operatorname{Grad}$ is defined with respect to the coordinates $\vec{X}$. Here, $\vec{x} \approx \vec{X}$ is assumed. Furthermore, $\deviatoricIndex{\ten{\epsilon}} = \ten{\epsilon} - \tr{\ten{\epsilon}} / 3 \ten{I}$ denotes the deviatoric part of $\ten{\epsilon}$. The constitutive model is parameterized in material parameters $\bm{\kappa} = \{K, G\}^\top$ composed of the bulk modulus $K$ and the shear modulus $G$.

\subsection{Hyperelasticity}\label{subsec:hyperelasticity}

In the following, we consider finite strains and compressible, isotropic hyperelastic materials. The first Piola-Kirchhoff stress tensor can be derived from a strain energy density function $\psiR$ expressed in terms of the tensor-valued measure $\ten{C}$ by
\begin{equation}\label{eq:model_hyperelasticity}
    \ten{P}(\ten{F}; \bm{\kappa}) = 2 \ten{F} \frac{\partial\psiR(\ten{C}; \bm{\kappa})}{\partial\ten{C}}.
\end{equation}
The deformation gradient $\ten{F}$ and the right Cauchy-Green tensor $\ten{C}$ are defined as
\begin{equation}\label{eq:deformation_gradient}
    \ten{F} = \frac{\partial\motionR(\vec{X},t)}{\partial\vec{X}} = \operatorname{Grad}{\vec{u}(\vec{X})} + \ten{I},
    \quad
    \ten{C} = \ten{F}^{\top} \ten{F},
\end{equation} 
where $\ten{I}$ is again the second-order identity tensor.

The strain energy density function $\psiR$ can be additively decomposed into a volumetric and an isochoric part $\psiVolR$ and $\psiIsoR$, respectively:
\begin{equation}\label{eq:strain_energy_split_hyperelasticity}
    \psiR(\ten{C}; \bm{\kappa}) = \psiVolR(\sca{J}; \bm{\kappa}) + \psiIsoR(\barten{C}; \bm{\kappa}).
\end{equation}
Here, $\sca{J} = \operatorname{det}(\ten{F})$ is the determinant of the deformation gradient and $\barten{C} = \sca{J}^{-2/3} \ten{C}$ is the isochoric right Cauchy-Green tensor. There are many concurrent approaches to model the volumetric part $\psiVolR$. A common approach frequently stated in standard text books \cite{holzapfel_nonlinearSolidMechanics_2000, wriggers_nonlinearFEM_2008} is to consider
\begin{equation}\label{eq:strain_energy_volumetric}
     \psiVolR(\sca{J}; \bm{\kappa}) = \frac{K}{4} ( \sca{J}^{2} - 1 - 2\operatorname{ln}\sca{J}),
\end{equation}
where $K$ is again the bulk modulus. For the isochoric part $\psiIsoR$,  a \textit{Neo-Hookean}-type ansatz
\begin{equation}\label{eq:strain_energy_isochoric}
    \psiIsoR(\barten{C}; \bm{\kappa}) = \frac{G}{2}(\sca{I}_{\barten{C}} - 3),
\end{equation}
with the first invariant $\sca{I}_{\barten{C}} = \operatorname{tr} (\barten{C})$ is chosen, where $G$ defines the shear modulus in the small strain limit. 

The relation between $K$ and $G$ might lead to a non-physical behavior for large compressive and tensile states, see, for a discussion, \cite{ehlers_largeVolumetricStrainsHyperelasticity_1998,hartmann_generalizedStrainEnergyFunctions_2003}. Thus, both the relation between $K$ and $G$ as well as the amount of the deformation has to be considered carefully. Again, as in the case of linear elasticity, the material parameters $\bm{\kappa}$ can be summarized as $\bm{\kappa} = \{K, G\}^\top$.


\section{Parametric physics-informed neural networks}\label{sec:PPINNs}

\Acfp{PINNs} are a deep learning framework for solving forward and inverse problems involving \acp{PDE}, in which \acp{ANN} act as a global ansatz function to the \ac{PDE} solution \cite{raissi_physicsInformedNeuralNetworks_2019}. An extension of the \ac{ANN} with additional inputs makes it even possible to learn parameterized forward solutions of \acp{PDE}. We first review the basics of \acp{ANN}. Subsequently, we lay emphasize on the key characteristic of parametric \acp{PINN} which is the formulation of the loss function. We further elaborate on necessary normalization steps for an application of the proposed parametric \ac{PINN} formulation in a real-world setting.

\subsection{Artificial neural networks}\label{subsec:ANNs}

\Acfp{ANN} are parameterized, nonlinear function compositions which serve as an approximation for an input-output mapping. There are several different formulations of this mapping, such as convolutional and recurrent neural networks. In the following, however, we restrict ourselves to fully-connected \acp{FFNN}. For a more in-depth introduction to \acp{ANN}, the reader is referred to standard text books, e.g., \cite{goodfellow_deepLearning_2016}.

We consider a fully-connected \ac{FFNN} $\ffnn$ composed of $\numLayers+1$ layers $\vec{h}^{(l)}$ that defines a mapping from an input space $\mathbb{R}^{N}$ to an output space $\mathbb{R}^{M}$ in the general form
\begin{equation}\label{eq:ffnn_definition}
\begin{aligned}
    \ffnn: \mathbb{R}^{N} &\to \mathbb{R}^{M}, \\
   \hat{\vec{x}} &\mapsto \ffnn(\hat{\vec{x}}) =  \bigl(\vec{h}^{(\numLayers)} \circ \vec{h}^{(\numLayers-1)} \circ \ldots \circ \vec{h}^{(1)}\bigr) \bigl(\hat{\vec{x}}\bigr),
\end{aligned}
\end{equation}
where $\hat{\vec{x}} \elm{N}$ denotes the input vector, $\hat{\vec{y}} \elm{M}$ the output vector and $\circ$ the composition operator, such that $(f \circ g)(x) = f(g(x))$). Accordingly, the first layer $\vec{h}^{(0)}$ and the last layer $\vec{h}^{(\numLayers)}$ are the input and the output layer, respectively, and defined as
\begin{equation}\label{eq:ffnn_input_output_layers}
    \vec{h}^{(0)} = \hat{\vec{x}} \elm{N}, \;\;
    \vec{h}^{(\numLayers)} = \hat{\vec{y}} \elm{M}.
\end{equation}
The $\numLayers - 1$ layers between the input and the output layer are usually called \textit{hidden layers}. The vector-valued output of the hidden layers and the output layer are defined as
\begin{equation}\label{eq:ffnn_hidden_output_layers}
    \vec{h}^{(l)} = \phi^{(l)}\Bigl(\vec{W}^{(l)}\vec{h}^{(l-1)} + \vec{b}^{(l)}\Bigr) = \phi^{(l)}\Bigl(\vec{z}^{(l)}\Bigr), \;\; l = \{1, \ldots, \numLayers\}.
\end{equation}
Here, $\vec{z}^{(l)}$ denotes the result of an affine transformation of the output vector of the downstream layer $\vec{h}^{(l-1)}$ controlled by the matrix $\vec{W}^{(l)}$ and the bias vector $\vec{b}^{(l)}$.
The output of the hidden layers is computed by applying a nonlinear \textit{activation function} $\phi^{(l)}$ on top of the affine transformation $\vec{z}^{(l)}$. In the output layer $\vec{h}^{(\numLayers)}$, the identity function is used as activation, such that
\begin{equation}\label{eq:ffnn_output_layer}
    \vec{h}^{(\numLayers)} = \phi^{(\numLayers)}\Bigl(\vec{W}^{(\numLayers)}\vec{h}^{(\numLayers-1)} + \vec{b}^{(\numLayers)}\Bigr) = \ten{I} \, \vec{z}^{(\numLayers)} = \vec{z}^{(\numLayers)},
\end{equation}
where $\ten{I}$ is the identity matrix of size $\numNeurons^{(\numLayers)} \times \numNeurons^{(\numLayers)}$ and $\numNeurons^{(\numLayers)}$ is the size of the vector $\vec{z}^{(\numLayers)}$ which is equivalent to the number of \textit{neurons} in this layer. In this contribution, we use the hyperbolic tangent as activation functions in the hidden layers.

The weight matrices $\vec{W}^{(l)}$ and bias vectors $\vec{b}^{(l)}$ comprise the trainable parameters of the layers $l = \{1, \ldots, \numLayers\}$. All parameters of the \ac{FFNN} can be combined in a single parameter vector $\bm{\theta}$ with
\begin{equation}\label{eq:ffnn_parameters}
    \bm{\theta} = \Bigl\{\vec{W}^{(l)},  \vec{b}^{(l)}\Bigr\}_{1 \leq l \leq \numLayers}.
\end{equation}
Taking the trainable parameters $\bm{\theta}$ into account, in the following, the \ac{FFNN} defined by \cref{eq:ffnn_definition,eq:ffnn_input_output_layers,eq:ffnn_hidden_output_layers,eq:ffnn_output_layer,eq:ffnn_parameters} is denoted by $\ffnn(\hat{\vec{x}}; \bm{\theta})$. This notation highlights that the \ac{FFNN} output $\hat{\vec{y}}$ does not only depend on the input $\hat{\vec{x}}$ but is also parameterized in the current realization of $\bm{\theta}$.

An appropriate point estimate of the parameters $\bm{\theta}$ can be found by solving an optimization problem, often referred to as \textit{training} of the \ac{ANN}. The objective of the optimization problem is to minimize a loss function that provides a measure for the deviation of the \ac{ANN} from the hidden input-output mapping. According to the universal function approximation theorem, any Borel measurable function can be approximated by an \ac{ANN} with enough parameters with only mild assumptions on the activation function \cite{cybenko_approximationBySuperpositions_1989, hornik_FFNUniversalApproximators_1989,li_approximationOfFunctionsByNN_1996}. However, it should be noted that the issue of finding the optimal parameters of the \ac{ANN} is still an open question and highly problem dependent.

\subsection{Parametric physics-informed neural network formulation}\label{subsec:PINNs_formulation}

Parametric \acp{PINN} are an extension of standard \acp{PINN} for learning parameterized forward solutions involving parametric \acp{PDE}. A parameterized ansatz is used to approximate the solution which is realized by an \ac{ANN} with additional inputs besides the spatial coordinates. In the following, we apply parametric \acp{PINN} for solving the model \cref{eq:balance_momentum,eq:bcs} parameterized in the material parameters $\bm{\kappa}$. We start by defining our ansatz function for the displacement field and the resulting discretized model. Subsequently, we formulate the loss function and elaborate on the training process.\\

\noindent \textbf{First, we approximate the displacement field by the parametric ansatz}
\begin{equation}\label{eq:ppinn_ansatz_standard}
    \vec{u}(\vec{X}, \bm{\kappa}) \approx \mathcal{U}(\vec{X}, \bm{\kappa}; \bm{\theta}),
\end{equation}
which acts as a function approximator to the solution of \cref{eq:balance_momentum,eq:bcs}. Here, $\mathcal{U}$ is a modified \ac{FFNN} $\ffnn$, whereby the modifications are explained later on. It should be noted that both the spatial coordinates $\vec{X}$ and the material parameters $\bm{\kappa}$ are inputs to the ansatz $\mathcal{U}$. The \ac{FFNN} is parameterized by the weights and biases $\bm{\theta}$ as defined in \cref{eq:ffnn_parameters}. Furthermore, in this work, we consider the calibration from full-field displacement data as a two-dimensional problem and thus $\mathcal{U}: \mathbb{R}^{2+\numKappa} \to \mathbb{R}^{2}$ where $\numKappa$ is the number of material parameters.

In particular, we use an ansatz for the displacement field that differs from a standard \ac{FFNN} as follows: We choose an ansatz function that strictly fulfills the Dirichlet boundary conditions \cref{eq:bcs_dirichlet} by construction, which is referred to as hard boundary conditions. Alternatively, the Dirichlet boundary conditions can be imposed by a separate loss term. This approach is referred to as soft boundary conditions. With the application of the hard boundary condition according to \cite{berg_unifiedANNApproachForPDEs_2018}, the \ac{FFNN} $\ffnn$ modifies to
\begin{equation}\label{eq:ppinn_ansatz_hard_bc}
    \intHBCAnsatzU(\vec{X}, \bm{\kappa}; \bm{\theta}) = \vec{G}(\vec{X}) + \vec{D}(\vec{X}) \otimes \ffnn(\barvec{X}; \bm{\theta}),
\end{equation}
where $\intHBCAnsatzU$ denotes an intermediate step in the derivation of the parameterized ansatz $\mathcal{U}$. Moreover, $\vec{G}$ is a extension of the boundary data with appropriate regularity and $\vec{D}$ is a smooth distance function giving the distance of $\vec{X} \in \bodyR$ to the boundary $\gammaDirichletR$. The vector $\barvec{X} = \{\vec{X}^{\top}, \bm{\kappa}^{\top} \}^{\top}$ is the summarized \ac{FFNN} input vector. When selecting the distance function, it is important to ensure that $\vec{D}$ vanishes on the boundary $\gammaDirichletR$. It should be noted that $\vec{G}$ and $\vec{D}$ are vector valued functions of the same dimension as the ansatz output and that $\otimes$ in \cref{eq:ppinn_ansatz_hard_bc} denotes the element-wise Hadamard multiplication operator, such that $[\vec{a} \otimes \vec{b}]_{i} = a_{i} \cdot b_{i}$ for two vectors $\vec{a}, \vec{b} \elm{n}$. In this contribution, we use a normalized linear distance function defined as
\begin{equation}\label{eq:ppinn_ansatz_distance_function}
    \vec{D}(\vec{X}) = (\vec{X} - \vec{X}_{\textrm{bc}}) \oslash (\maxIndex{\vec{X}} - \minIndex{\vec{X}}),
\end{equation}
where $\minIndex{\vec{X}}$ and $\maxIndex{\vec{X}}$ are vectors containing the minimum and maximum coordinates for each dimension within $\bodyR$, respectively. In addition, $\vec{X}_{\textrm{bc}}$ is a vector that contains the position of the Dirichlet boundary condition in the respective dimension. The element-wise Hadamard division operator $\oslash$ is defined as $[\vec{a} \oslash \vec{b}]_{i} = a_{i} / b_{i}$ for two vectors $\vec{a}, \vec{b} \elm{n}$. Note that the distance function defined in \cref{eq:ppinn_ansatz_distance_function} assumes that there is only one Dirichlet boundary condition in each dimension and that the Dirichlet boundaries are parallel to the Cartesian coordinate system. In general, however, hard boundary conditions can also be applied to complex geometries, as shown in \cite{berg_unifiedANNApproachForPDEs_2018}.

Furthermore, we normalize the inputs and outputs of the ansatz because it is well known that this accelerates the convergence of the training of \acp{ANN}. According to \cite{lecun_efficientBackProp_2012}, the mean value of each input feature should be close to zero. Since we assume that the input is evenly distributed over the input domain, we normalize the input by the following linear transformation
\begin{equation}\label{eq:normalization_ann_inputs}
    \vec{N}_{\ffnnInput}(\barvec{X}) = 2 (\barvec{X} - \minIndex{\barvec{X}}) \oslash (\maxIndex{\barvec{X}} - \minIndex{\barvec{X}}) - \bm{1},
\end{equation}
which maps the entries of the real input vector $\barvec{X}$ to the range $[-1, 1]$. Here, $\minIndex{\barvec{X}}$ and $\maxIndex{\barvec{X}}$ are vectors containing the minimum and maximum input features, respectively, and $\bm{1}$ is a vector of ones. In addition, we normalize the ansatz outputs. Depending on the problem, the scales of the displacements can vary significantly in the different dimensions, as, e.g., in uniaxial tensile tests. At the same time, error metrics like the mean squared error are scale-sensitive. To give the displacement field approximation the same relative importance in all dimensions during training, we enforce the ansatz outputs to be also in the range $[-1, 1]$. Therefore, we renormalize the output in a last step by another linear transformation
\begin{equation}\label{eq:renormalization_ansatz_output}
    \vec{N}^{-1}_{\ffnnOutput}\Bigl(\intNormAnsatzU(\vec{X}, \bm{\kappa}; \bm{\theta})\Bigr) = \frac{1}{2} \Bigl(\intNormAnsatzU(\vec{X}, \bm{\kappa}; \bm{\theta}) + \vec{1}\Bigr) \otimes (\maxIndex{\vec{u}} - \minIndex{\vec{u}}) + \minIndex{\vec{u}}, 
\end{equation}
where $\intNormAnsatzU$ is the intermediate normalized ansatz with its outputs enforced to be in the range $[-1, 1]$. The vectors $\minIndex{\vec{u}}$ and $\maxIndex{\vec{u}}$ contain the minimum and maximum expected displacements of the material body resulting from the range of material parameters $\bm{\kappa}$ under consideration, respectively. The intermediate normalized ansatz is defined as
\begin{equation}\label{eq:ppinn_ansatz_normalized}
    \intNormAnsatzU(\vec{X}, \bm{\kappa}; \bm{\theta})
    =  \vec{N}_{\ffnnOutput}\Bigl(\vec{G}(\vec{X})\Bigr) 
    + \vec{D}(\vec{X}) 
    \otimes \ffnn\Bigl(\vec{N}_{\ffnnInput}(\barvec{X}); \bm{\theta}\Bigr).
\end{equation}
In order to guarantee that the renormalized ansatz output $\mathcal{U}$ still strictly fulfills the Dirichlet boundary conditions, the boundary extension $\vec{G}$ in \cref{eq:ppinn_ansatz_normalized} must also be normalized by the inverse of \cref{eq:renormalization_ansatz_output} which is given by
\begin{equation}\label{eq:normalization_ansatz_output}
    \vec{N}_{\ffnnOutput}\Bigl(\vec{G}(\vec{X})\Bigr) = 2 \Bigl(\vec{G}(\vec{X}) - \minIndex{\vec{u}}\Bigr) \oslash (\maxIndex{\vec{u}} - \minIndex{\vec{u}}) - \bm{1}.
\end{equation}
Note that $\vec{D}(\vec{X})$ in \cref{eq:ppinn_ansatz_normalized} is also normalized by definition \cref{eq:ppinn_ansatz_distance_function}.

Applying the normalization and renormalization steps from equations \cref{eq:ppinn_ansatz_distance_function,eq:normalization_ann_inputs,eq:renormalization_ansatz_output,eq:ppinn_ansatz_normalized,eq:normalization_ansatz_output} to the modified ansatz $\intHBCAnsatzU$ from equation \cref{eq:ppinn_ansatz_hard_bc}, we finally obtain the ansatz
\begin{equation}\label{eq:ppinn_ansatz}
\begin{split}
    \mathcal{U}(\vec{X}, \bm{\kappa}; \bm{\theta}) &= \vec{N}^{-1}_{\ffnnOutput}\Bigl(\intNormAnsatzU(\vec{X}, \bm{\kappa}; \bm{\theta})\Bigr) \\
    &=  \vec{N}^{-1}_{\ffnnOutput}\biggl(
    \vec{N}_{\ffnnOutput}\Bigl(\vec{G}(\vec{X})\Bigr) 
    + \vec{D}(\vec{X}) 
    \otimes \ffnn\Bigl(\vec{N}_{\ffnnInput}(\barvec{X}); \bm{\theta}\Bigr)
    \biggr).
\end{split}
\end{equation}
The normalization steps aim to condition the optimization problem that arises during \ac{PINN} training. While the required minimum and maximum input values are given from the training data, the required minimum and maximum output values can, e.g., be extracted from given experimental or simulation data or be estimated based on prior knowledge such as boundary conditions. It is important to emphasize that at any time during training and prediction, only the non-normalized, extended inputs $\barvec{X}$ are fed into the ansatz. Likewise, the ansatz always outputs non-normalized displacements. This also means that the physics is not violated when the outputs are derived with respect to the inputs during training.

For the following steps, we reformulate the governing equations introduced in \cref{sec:mechanics} as a function of the displacement state vector $\uModelState$ and the material parameters $\bm{\kappa}$ and define the discretized model
\begin{equation}\label{eq:model}
    \vecF(\uModelState, \bm{\kappa}) = 
    \begin{Bmatrix*}[l]
    \vecFC(\uModelStateC, \bm{\kappa}) \\
    \vecFD(\uModelStateD, \bm{\kappa}) \\
    \vecFN(\uModelStateN, \bm{\kappa}) \\
    \end{Bmatrix*} = 
    \begin{Bmatrix*}[r]
        \operatorname{Div} \ten{P}(\uModelStateC; \bm{\kappa}) + \rohR \vec{b} \\
        \uModelStateD - \barvec{u}_{\textrm{F}} \\
        \ten{P}(\uModelStateN; \bm{\kappa}) \cdot \normalR - \barvec{t}_{\textrm{F}} \\
    \end{Bmatrix*}.
\end{equation}
A statically and kinematically admissible displacement field must fulfill $\vecF = \vec{0}$ everywhere in $\bodyR$. When training \acp{PINN} to solve the forward problem, we minimize a loss based on $\vecF(\uModelState, \bm{\kappa})$. The displacement, however, is only evaluated at discrete points, represented in the model state vector $\uModelState \elm{2(\numC + \numD + \numN)}$. The latter comprises the displacement state vectors $\uModelStateC \elm{2\numC}$, $\uModelStateD \elm{2\numD}$ and $\uModelStateN \elm{2\numN}$, where $\numC$, $\numD$ and $\numN$ are the number of evaluation points inside the domain $\bodyR$ and on the Dirichlet and Neumann boundaries $\gammaDirichletR$ and $\gammaNeumannR$, respectively. Accordingly, $\vecF \elm{2(\numC + \numD + \numN)}$ comprises $\vecFC \elm{2\numC}$, $\vecFD \elm{2\numD}$ and $\vecFN \elm{2\numN}$. Furthermore, $\barvec{u}_{\textrm{F}} \elm{2\numD}$ and $\barvec{t}_{\textrm{F}} \elm{2\numN}$ are the vectors with the prescribed displacements and tractions, respectively. The implementation of the discrete model \cref{eq:model} for solving the forward problem using \acp{PINN} is introduced in the following.\\

\noindent \textbf{Second, we define the loss function.} The loss function encoding the physics in the model \cref{eq:model} and enhanced by data is defined as
\begin{equation}\label{eq:ppinn_loss}
    \lossF(\bm{\theta}; \dataSet) 
    = \lambdaC \lossFC(\bm{\theta}; \dataSetC ) 
    + \lambdaN \lossFN(\bm{\theta}; \dataSetN ) 
    + \lambdad \lossFd(\bm{\theta}; \dataSetd ).
\end{equation}
The loss terms $\lossFC$, $\lossFN$ and $\lossFd$ penalize the mean squared error of the approximation $\mathcal{U}$ defined in \cref{eq:ppinn_ansatz} with respect to the \ac{PDE}, the Neumann boundary condition and the data, respectively, and are defined as
\begin{subequations}\label{eq:ppinn_loss_terms}
\begin{align}
    \lossFC(\bm{\theta}; \dataSetC) &= \frac{1}{2\numC} \sum_{i=1}^{\numC} \norm{ \vecFC\Bigl( {\uModelStateC}^{(i)}, \bm{\kappa}^{(i)} \Bigr) }^{2} \label{eq:ppinn_loss_terms_pde}\\
    &= \frac{1}{2\numC} \sum_{i=1}^{\numC} \norm{ \operatorname{Div} \ten{P}\Bigl(\mathcal{U}(\vec{X}^{(i)}, \bm{\kappa}^{(i)}; \bm{\theta}); \bm{\kappa}^{(i)} \Bigr) + \rohR\Bigl(\vec{X}^{(i)}\Bigr) \, \vec{b}\Bigl(\vec{X}^{(i)}\Bigr) }^{2}, \nonumber \\
    \lossFN(\bm{\theta}; \dataSetN) &= \frac{1}{2\numN} \sum_{k=1}^{\numN} \norm{ \vecFN\Bigl( {\uModelStateN}^{(k)}, \bm{\kappa}^{(k)} \Bigr) }^{2} \label{eq:ppinn_loss_terms_neumannbc}\\
    &= \frac{1}{2\numN} \sum_{k=1}^{\numN} \norm{ \ten{P}\Bigl(\mathcal{U}(\vec{X}^{(k)}, \bm{\kappa}^{(k)}; \bm{\theta}); \bm{\kappa}^{(k)} \Bigr) \cdot \normalR\Bigl(\vec{X}^{(k)}\Bigr) - \barvec{t}_{\textrm{F}}^{(k)} }^{2}, \nonumber \\
    \lossFd(\bm{\theta}; \dataSetd ) &= \frac{1}{2\numd} \sum_{l=1}^{\numd} \norm{ \mathcal{U}(\vec{X}^{(l)}, \bm{\kappa}^{(l)}; \bm{\theta}) - \barvec{u}_{\textrm{d}}^{(l)} }^{2}, \label{eq:ppinn_loss_terms_data}
\end{align}
\end{subequations}
where $\norm{\bullet}^{2}$ denotes the squared $\LTwo$-norm. The training data $\dataSet$ consists of three sets $\dataSetC$, $\dataSetN$ and $\dataSetd$:
\begin{enumerate}[label=(\roman*)]
    \item $\dataSetC$ is referred to as a set of $\numC$ collocation points $\left\{\vec{X}^{(i)}, \bm{\kappa}^{(i)} \right\}_{i=1}^{\numC}$ sampled from the domain $\bodyR$.
    \item $\dataSetN$ consists of $\numN$ collocation points $\left\{\vec{X}^{(k)}, \bm{\kappa}^{(k)}, \barvec{t}_{\textrm{F}}^{(k)} \right\}_{k=1}^{\numN}$ on the Neumann boundary $\gammaNeumannR$ with the prescribed tractions $\barvec{t}_{\textrm{F}}^{(k)}$. 
    \item $\dataSetd$ contains $\numd$ points $\left\{\vec{X}^{(l)}, \bm{\kappa}^{(l)}, \barvec{u}_{\textrm{d}}^{(l)} \right\}_{l=1}^{\numd}$ where the displacements $\barvec{u}_{\textrm{d}}^{(l)}$ can be obtained from, e.g., \ac{FE} simulations. 
\end{enumerate}
The individual loss terms in \cref{eq:ppinn_loss} can additionally be weighted by $\lambdaC$, $\lambdaN$ and $\lambdad$ to balance them. The weight factors may also be adapted during training, see, for instance, \cite{wang_gradientFlowPathologiesInPINNs_2021}. In order to calculate the partial derivatives required to evaluate the loss terms \cref{eq:ppinn_loss_terms_pde,eq:ppinn_loss_terms_neumannbc}, the displacement field in the constitutive models \cref{eq:model_linearelasticity} and \cref{eq:model_hyperelasticity} is approximated by the ansatz \cref{eq:ppinn_ansatz_standard}. The derivatives of the ansatz outputs with respect to the inputs are calculated using automatic differentiation \cite{baydin_automaticDifferentiation_2018}. If required, the loss function \cref{eq:ppinn_loss} may be complemented by further, problem specific loss terms, such as symmetry boundary conditions.

It should be noted that the loss function \cref{eq:ppinn_loss} does not contain a separate loss term for the Dirichlet boundary condition since we use a hard boundary condition for this, see \cref{eq:ppinn_ansatz}. Provided that the stress is also considered as an output of the \ac{ANN} in addition to the displacement, the Neumann boundary condition can in principle also be replaced by a hard boundary condition. In this work, we do not use hard Neumann boundary conditions, as we achieved high accuracy without them and do not observe any problems with the weak imposition of the Neumann boundary conditions.\\

\noindent \textbf{Third, we optimize the \ac{ANN} parameters $\bm{\theta}$}. The optimization problem for finding an appropriate point estimate for the \ac{ANN} parameters $\bm{\theta}^{*}$ is defined as 
\begin{equation}\label{eq:ppinn_optimization_problem}
    \bm{\theta}^{*} = \argmin_{\bm{\theta}} \lossF(\bm{\theta}; \dataSet),
\end{equation}
and is usually carried out using gradient-based optimization algorithms, such as ADAM \cite{kingma_ADAM_2015} or L-BFGS \cite{liu_LBFGS_1989,broyden_BFGS_1970,fletcher_BFGS_1970,goldfarb_BFGS_1970,shanno_BFGS_1970}. The required gradients of the loss function $\lossF$ with respect to the \ac{ANN} parameters $\bm{\theta}$ can again be calculated by automatic differentiation. It should be noted, that the implementation of $\lossF$ in \cref{eq:ppinn_loss} is not identical to the model formulation \cref{eq:model}. However, $\lossF = 0$ implies that $\vecF = \vec{0}$. Squaring the residuals in $\lossF$ ensures that positive and negative deviations do not cancel out each other. In addition, larger residuals are penalized more than smaller residuals.


\section{Constitutive model calibration}\label{sec:problem_statement}
 
In this contribution, we formulate the calibration from full-field displacement data according to the reduced approach. Following the problem statement, we elaborate on the deterministic nonlinear least-squares method. Afterwards, we address the calibration problem from a Bayesian statistical point of view. In both the deterministic as well as the Bayesian statistical setting, the parametric \ac{PINN} is used as a surrogate for the mechanical model.

\subsection{Deterministic calibration}\label{subsec:problem_statement_deterministic}
 
\textbf{Problem statement:} Recalling the notation from \cref{sec:mechanics,sec:PPINNs}, constitutive model calibration is based on the following equation
\begin{equation}\label{eq:abstract_calibration}
    \vec{O}(\uState) = \vec{d},
\end{equation}
with state vector $\uState \in \Omega_{\vec{u}} \subset \mathbb{R}^{2\numu}$ calculated by the pre-trained parametric \ac{PINN}, full-field displacement data $\vec{d} \elm{2\numd}$ and observation map $\vec{O}$. 
The latter relates the model state $\uState$ to the measurement data $\vec{d}$, such that $\vec{O}(\uState) \elm{2\numd}$. In principle, the observation map can take many forms and may also account for indirectly measured quantities, such as strains. If full-field displacement measurements are available, it interpolates the model state $\uState$ to the $\numd$ sensor locations $\left\{\vec{X}^{(m)}\right\}^{\numd}_{m=1}$. These are the points where the displacement is measured. It is worth recalling that the \ac{PINN} is a global ansatz function that can be evaluated directly at the sensor locations. Consequently, the observation map becomes the identity operator, i.e., $\vec{O}(\uState)= \ten{I} \, \uState= \uState$, where $\ten{I}$ is the identity matrix of size $2\numu \times 2\numu$ and $\numu = \numd$. Hence, possible interpolation errors are avoided. \\

\noindent \textbf{Solution approach:} In the reduced formulation, the implicit function theorem is applied, see, e.g., \cite{krantz_implicitFunctionTheorem_2013}, and the state vector is expressed directly as a function of the parameters via $\uState = \hatUState(\bm{\kappa})$, where $\bm{\kappa} \in \Omega_{\bm{\kappa}} \subset \mathbb{R}^{\numKappa}$ is the material parameter vector. Accordingly, the displacement at a material point $\vec{X}$ is expressed via $\vec{u}(\vec{X}) = \hatvec{u}(\vec{X},\bm{\kappa})$.

The parameters-to-state map, also known as solution map, is here provided by the pre-trained \ac{PINN} $\mathcal{U}$. The state vector is defined as
\begin{equation}\label{eq:parameters_to_state_map}
    \hatUState(\bm{\kappa}) = 
    \begin{Bmatrix*}
        \hatsca{u}_{x}(\vec{X}^{(1)},\bm{\kappa}) \\
        \vdots \\
        \hatsca{u}_{x}(\vec{X}^{(m)},\bm{\kappa}) \\
        \hatsca{u}_{y}(\vec{X}^{(1)},\bm{\kappa}) \\
        \vdots \\
        \hatsca{u}_{y}(\vec{X}^{(m)},\bm{\kappa})
    \end{Bmatrix*} =
    \begin{Bmatrix*}
        \mathcal{U}_{x}(\vec{X}^{(1)}, \bm{\kappa}; \bm{\theta}) \\
        \vdots \\
        \mathcal{U}_{x}(\vec{X}^{(m)}, \bm{\kappa}; \bm{\theta}) \\
        \mathcal{U}_{y}(\vec{X}^{(1)}, \bm{\kappa}; \bm{\theta}) \\
        \vdots \\
        \mathcal{U}_{y}(\vec{X}^{(m)}, \bm{\kappa}; \bm{\theta})
    \end{Bmatrix*},
\end{equation}
where the subscript in $\hatsca{u}_{\bullet}$ and $\mathcal{U}_{\bullet}$ denotes the dimension. With the parameters-to-state map defined in \cref{eq:parameters_to_state_map}, we can recast the data model as follows
\begin{equation}\label{eq:deterministic_reduced_observation}
    \hatUState(\bm{\kappa}) = \vec{d}.
\end{equation}
The parameters-to-state map $\hatUState(\bm{\kappa})$ is obtained by pre-training the \ac{PINN} $\mathcal{U}$ prior to the calibration for the parameter set $\Omega_{\bm{\kappa}}$. After pre-training, the \ac{ANN} parameters $\bm{\theta}$ are frozen. Thus, in an online stage, the constitutive model can be calibrated solely on \cref{eq:deterministic_reduced_observation}. The main advantage of this formulation is that the resulting optimization or inference problem only needs to be solved in the parameter domain $\Omega_{\bm{\kappa}}$. 

The deterministic, reduced calibration problem stated in \cref{eq:deterministic_reduced_observation} can be reformulated as a \acf{NLS} optimization problem. Therefore, \cref{eq:deterministic_reduced_observation} is rearranged to define the residual $\vec{r}$ as
\begin{equation}\label{eq:nls_residuals}
    \vec{r}(\bm{\kappa}) = \hatUState(\bm{\kappa}) - \vec{d}.
\end{equation}
In order to account for different magnitudes of the displacements in each dimension, we consider weighted residuals $\tilde{\vec{r}}(\bm{\kappa}) = \vec{W} \, \vec{r}(\bm{\kappa})$ with the diagonal weight matrix $\vec{W} \elmm{2\numd}{2\numd}$, see \cite{lawson_leastSquaresProblems_1995}. Especially in the context of parameter identification, a weight matrix can also be introduced to take into account different physical quantities or a meaningful scaling of observations that are not all equally reliable \cite{mahnken_identificationOfMaterialParameters_2017}. The weight matrix is assembled as
\begin{equation}\label{eq:nls_weight_matrix_assembling}
    \vec{W} := \begin{bmatrix}
        \vec{W}_{x} & \vec{0} \\
        \vec{0} & \vec{W}_{y}
    \end{bmatrix}, \; \vec{W} \elmm{2\numd}{2\numd},
\end{equation}
where we define the sub-weight matrices $\vec{W}_{x}, \vec{W}_{y} \elmm{\numd}{\numd}$ as
\begin{equation}\label{eq:nls_weight_matrix_submatrices}
    \vec{W}_{x} = \frac{1}{\meanSup{\sca{u}_{x}}} \ten{I} \;\; \text{and} \;\; \vec{W}_{y} = \frac{1}{\meanSup{\sca{u}_{y}}} \ten{I}, 
\end{equation}
with the identity matrix $\ten{I}$ of size $\numd \times \numd$ and the mean absolute displacements $\meanSup{\sca{u}_{x}}$ and $\meanSup{\sca{u}_{y}}$ in $x$- and $y$-direction determined as
\begin{equation}\label{eq:nls_weight_matrix_meandispalcements}
     \meanSup{\sca{u}_{x}} = \frac{1}{\numd} \sum_{i=1}^{\numd} \vert \sca{u}_{x}^{(i)}\vert \;\; \text{and} \;\; \meanSup{\sca{u}_{y}} = \frac{1}{\numd} \sum_{i=1}^{\numd} \vert \sca{u}_{y}^{(i)}\vert.
\end{equation}
The loss function $\phi(\bm{\kappa})$ is then given by the sum of the squared, weighted residuals as
\begin{equation}\label{eq:nls_loss}
    \phi(\bm{\kappa}) = \half \norm{\tilde{\vec{r}} (\bm{\kappa})}^{2} = \half \norm{\vec{W} \, (\hatUState(\bm{\kappa}) - \vec{d})}^{2}.
\end{equation}
A deterministic point estimate of the material parameters $\bm{\kappa}^{*}$ can be determined by solving the minimization problem
\begin{equation}\label{eq:nls_optimization_problem}
    \bm{\kappa}^{*} = \argmin_{\bm{\kappa}} \phi(\bm{\kappa}) \text{ subject to } \bm{\kappa} \in \Omega_{\bm{\kappa}},
\end{equation}
where $\bm{\kappa}^{*}$ must be a value from the set $\Omega_{\bm{\kappa}}$ which contains only physically admissible material parameters. The so-called normal equation is recovered from the necessary condition of a vanishing gradient of the loss function $\phi(\bm{\kappa})$ at the solution $\bm{\kappa}^{*}$,
\begin{equation}
    \label{eq:nls_normal_equation}
    \left.\frac{\d{\phi(\bm{\kappa})}}{\d{\bm{\kappa}}}\right|_{\bm{\kappa}=\bm{\kappa}^*} = \left[\frac{\d{\hatUState(\bm{\kappa}^{*})}}{\d{\bm{\kappa}}}\right]^\top 
    \vec{W}^\top \vec{W} \, (\hatUState(\bm{\kappa}^*)-\vec{d}) = \vec{0},
\end{equation}
which is in general a system of nonlinear equations. Here, $\d{\hatUState(\bm{\kappa}^{*})}/\d{\bm{\kappa}} \elmm{2\numd}{\numKappa}$ is the Jacobian of the parameters-to-state map $\hatUState$ with respect to the material parameters $\bm{\kappa}$ and can be calculated with automatic differentiation when using \acp{PINN}.

Problem \cref{eq:nls_optimization_problem} can be solved using well-established optimization procedures, such as gradient-based or gradient-free techniques. In particular, we use the L-BFGS algorithm. It should be noted that multiple global or local minima of problem \cref{eq:nls_optimization_problem} may exist. In this case, $\bm{\kappa}^{*}$ is an arbitrary element of the solution set of the minimization problem that depends, among others, on the initial material parameter values. This leads to the concept of \textit{local identifiability} of material parameters and is addressed in \cite{roemer_modelCalibrationInSolidMechanics_2024} when using full-field data.

\subsection{Bayesian statistical inference}\label{subsec:problem_statement_statistical}

\textbf{Problem statement:} Constitutive model calibration can also be addressed from a Bayesian statistical point of view. In this setting, the unknown material parameters are treated as random variables with prior probability distributions $p(\bm{\kappa})$. The prior distribution is then updated according to Bayes' law
\begin{equation}\label{eq:bayes_law}
    p(\bm{\kappa} \vert \vec{d}) \propto p(\vec{d} \vert \bm{\kappa}) p(\bm{\kappa}), 
\end{equation}
where $p(\bm{\kappa} \vert \vec{d})$ is the posterior probability density and $p(\vec{d} \vert \bm{\kappa})$ represents the likelihood function \cite{gelman_bayesianDataAnalysis_2013}.
In analogy to the deterministic formulation defined in \cref{eq:deterministic_reduced_observation}, the statistical counterpart reads
\begin{equation}\label{eq:bayesian_reduced_observation}
\hatUState(\bm{\kappa}) = \vec{d} + \vec{e},
\end{equation}
with the observation noise vector $\vec{e} \sim \mathcal{N}(\vec{0}, \bm{\Sigma}_{\vec{e}})$. We assume that the noise $\vec{e}$ in the measurement data is normally distributed with zero mean and positive definite covariance matrix $\bm{\Sigma}_{\vec{e}}$. In addition, we assume the noise to be independent and identically distributed (i.i.d), leading to a diagonal covariance matrix with entries $\sigma_{e}^2$.\\

\noindent \textbf{Solution approach:} Under the aforementioned assumptions, equation \cref{eq:bayesian_reduced_observation} implies the conditional probability density
\begin{equation}\label{eq:likelihood_function}
\begin{aligned}
    p(\vec{d} \vert \bm{\kappa}) &= \mathcal{N}(\hatUState(\bm{\kappa}),\bm{\Sigma}_{\vec{e}}) \\
    &=  \frac{1}{(2 \pi)^{2\numd/2} \det{\bm{\Sigma}_{\vec{e}}}^{1/2}} 
    \mathrm{exp}\Bigl(-\frac{1}{2} (\hatUState(\bm{\kappa}) - \vec{d})^\top
    \bm{\Sigma}_{\vec{e}}^{-1} 
    (\hatUState(\bm{\kappa}) - \vec{d})\Bigr),
\end{aligned}
\end{equation}
corresponding to the likelihood function of the data $\likelihood(\bm{\kappa}) := p(\vec{d} \vert \bm{\kappa})$. The likelihood function expresses the plausibility of observing the data $\vec{d}$ for given material parameters $\bm{\kappa}$. 

It is important to note that even for normally distributed data and a normally distributed prior, the posterior is not normally distributed for finite sample sizes. This is because of a nonlinear dependence between the material parameters $\bm{\kappa}$ and the simulated displacements $\hatUState$. These nonlinearities may be introduced directly by a nonlinear constitutive model. But even in the case of linear constitutive models, the map from material parameters to displacements is nonlinear, only the forcing influences the displacements in a linear way. This means that the posterior cannot be determined analytically in a closed-form expression. Instead, the posterior can be approximated numerically by a sampling-based \acf{MCMC} analysis. In our numerical tests, we use a stable and well-tested implementation of the affine-invariant ensemble sampler, also known as \texttt{emcee} \cite{goodman_ensembleSamplersAffineInvariance_2010}. This algorithm is robust and in comparison to other \ac{MCMC} algorithms, it does require hand-tuning of only one hyperparameter, which is the stretch scale. For an in-depth description of the algorithm behind \texttt{emcee} and an explanation of the hyperparameter, please refer to \cite{foremanMackey_emcee_2013}.

Once the posterior distribution is determined, it provides both a point estimate as well as a quantification of uncertainty. The maximum a posteriori estimate is given by
\begin{equation}\label{eq:maximum_a_posteriori_estimate}
\begin{aligned}
    \bm{\kappa}^{*} &= \argmin_{\bm{\kappa}} -\log p(\bm{\kappa} \vert \vec{d} ) \\
    &= \argmin_{\bm{\kappa}} - \bigl( \log \likelihood(\bm{\kappa}) + \log p(\bm{\kappa}) \big).
\end{aligned}
\end{equation}
Substituting the likelihood function $\likelihood(\bm{\kappa})$ from \cref{eq:likelihood_function}, we obtain 
\begin{equation}\label{eq:maximum_a_posteriori_estimate_likelihood_substituted}
    \bm{\kappa}^{*} = \argmin_{\bm{\kappa}} \Bigl( \frac{1}{2} \norm{ \hatUState(\bm{\kappa}) - \vec{d} }_{\bm{\Sigma}_{\vec{e}}^{-1}}^{2} - \log p(\bm{\kappa}) \Bigr),
\end{equation}
with the weighted norm $\norm{\vec{b}}_{\ten{A}}^2 = \vec{b}^{\top} \ten{A} \vec{b}$  for any positive definite matrix $\vec{A}$.
For a Gaussian prior, the maximum a posteriori estimate naturally leads to a regularized \ac{NLS} problem.\\

\noindent \textbf{Uncertainty quantification from a \textit{frequentist perspective}:} The uncertainty of a point estimate can be quantified through credible intervals which can be derived on the basis of the posterior distribution, and are also referred to as posterior intervals \cite{gelman_bayesianDataAnalysis_2013}. A credible interval is associated with an interval in the parameter domain, containing an unknown parameter $\kappa_{i}$ with a certain probability. Provided that the posterior probability density of the parameter $\kappa_{i}$ is normally distributed, such that $p(\kappa_{i} \vert \vec{d}) \approx \mathcal{N}(\mu_{p(\kappa_{i} \vert \vec{d})}, \sigma_{p(\kappa_{i} \vert \vec{d})} )$, the unknown $\kappa_{i}$ has a value in the credible interval $\textrm{CI}_{\textrm{95\%}} = \left[\mu_{p(\kappa_{i} \vert \vec{d})} \pm 1.96 \cdot \sigma_{p(\kappa_{i} \vert \vec{d})}\right]$ with a probability of approximately $\SI{95}{\percent}$. 

In a \textit{Bayesian setting}, a correct uncertainty quantification relies on an accurate parameters-to-state map. However, if the parameters-to-state map is misspecified, e.g., by simplifying modeling assumptions or simply by numerical errors, it follows that the model $\vecF \neq \vec{0}$. This also leads to a misspecified statistical model by the likelihood function $\likelihood(\bm{\kappa})$. As a consequence, the quantification of uncertainty may not be valid \cite{kleijn_bernsteinVonMisesUnderMisspecification_2012}. The correctness and validity of the uncertainty quantification must therefore be verified. As illustrated above, from a frequentist point of view, the uncertainty is valid if for $\numTests \rightarrow \infty$ experiments the true material parameter has probability $\alpha$ to be within the credible interval $\textrm{CI}_{\alpha}$, i.e., if the credible intervals are also confidence intervals. The validity of the uncertainty quantification from a frequentist perspective can thus be determined by performing a \textit{coverage test}. The coverage test can be used to assess how well the credible interval covers the true parameter and is described below in more detail. First, the posterior distribution $p(\bm{\kappa} \vert \vec{d})$ is determined for a large number of independent tests $\numTests$. Second, the frequency $\beta^{(i)}=n_{\textrm{CI}_{\alpha}}^{(i)}/\numTests$ of the true parameter $\kappa_{i}$ to be within the credible interval $\textrm{CI}_{\alpha}^{(i)}$ is calculated. Here, $n_{\textrm{CI}_{\alpha}}^{(i)}$ is the number of tests for which $\kappa_{i} \in \textrm{CI}_{\alpha}^{(i)}$. Note that the coverage $\beta^{(i)}$ is calculated separately for each parameter $\kappa_{i}$ for simplicity. Since the true parameters $\bm{\kappa}$ must be known for the test, we use synthetic data for which the parameters are then re-identified. Finally, the estimated uncertainty for parameter $\kappa_{i}$ is valid if $\beta^{(i)} \approx \alpha$.


\section{Results for synthetic full-field data}\label{sec:results_synthetic}

In the following, we demonstrate the calibration of constitutive models from synthetic full-field displacement data using parametric \acp{PINN}. Both small strain linear elasticity and finite strain hyperelasticity are considered. First, we define the test cases and the hyperparameters of both the parametric \acp{PINN}' architecture and the training settings. We then start with the deterministic calibration by solving the \ac{NLS} problem. We further quantify the uncertainty in the estimated material parameters by conducting Bayesian statistical inference. All results are statistically analyzed.

\subsection{Test cases and training of parametric PINNs}\label{subsec:testcases_synthetic}

\begin{figure}
    \centering
    \begin{tikzpicture}[scale=0.8, node distance=3.5cm]
        \tikzstyle{offlinestep} = [
            rectangle, 
            rounded corners, 
            minimum width=2.5cm, 
            minimum height=1.2cm,
            text centered, 
            draw=black, 
            fill=purple!50!gray!20
        ]
        \tikzstyle{onlinestep} = [
            rectangle, 
            rounded corners, 
            minimum width=2.5cm, 
            minimum height=1.2cm,
            text centered, 
            draw=black, 
            fill=green!40!gray!30
        ]
        \tikzstyle{arrow} = [thick,->,>=stealth, rounded corners]
        \tikzstyle{timeline} = [thick,->,>=stealth, width=1mm]

        \newcommand\xshiftcalibration{1.0cm};
        \newcommand\yshiftcalibration{1.2cm};
        
        \node (datageneration) [
        offlinestep, align=center
        ] {\textbf{data}\\ \textbf{generation}\\ using \ac{FEM}};
        
        \node (training) [
        offlinestep, right of=datageneration, align=center
        ] {\textbf{training}\\ of parametric\\ \ac{PINN}};
        
        \node (validation) [
        offlinestep, right of=training, align=center
        ] {\textbf{validation}\\ of parametric\\ \ac{PINN}};
        
        \node (calibrationdet) [
        onlinestep, above right of=validation, align=center, xshift=\xshiftcalibration, yshift=-\yshiftcalibration
        ] {deterministic\\ \textbf{calibration}};
        
        \node (calibrationstat) [
        onlinestep, below right of = validation, align=center, xshift=\xshiftcalibration, yshift=\yshiftcalibration
        ] {statistical\\ \textbf{calibration}};

        \draw [arrow] (datageneration) -- (training);
        \draw [arrow] (datageneration) -- +(0,1.5cm) -| ([shift=({-1cm,0})]validation);
        \draw [arrow] (training) -- (validation);
        \draw [arrow] (validation) |- node[shift=({1cm,0.25cm})] {\ac{NLS}} (calibrationdet);
        \draw [arrow] (validation) |- node[shift=({1cm,-0.25cm})] {\ac{MCMC}} (calibrationstat);

        \newcommand\xshifttimeline{1.5cm};
        \newcommand\yshifttimeline{-3cm};
        \newcommand\linewidthtimeline{0.5mm};
        \node (ghostnodstoponline) [right of = validation] {};
        \coordinate (startoffline) at ([shift=({-\xshifttimeline,\yshifttimeline})]datageneration);
        \coordinate (startonline) at ([shift=({\xshifttimeline,\yshifttimeline})]validation);
        \coordinate (stoponline) at ([shift=({\xshifttimeline,\yshifttimeline})]ghostnodstoponline);
        \draw [|-|, line width=\linewidthtimeline] (startoffline) -- (startonline) node[midway, below] {offline stage};
        \draw [->, line width=\linewidthtimeline] (startonline) -- (stoponline) node[midway, below] {online stage};
        
    \end{tikzpicture}
    \caption{Flowchart of the entire process including the offline as well as the online stage. In the offline stage, the data for both training and validation is generated using \ac{FEM}. The parametric \ac{PINN} is then trained and validated. In the online stage, the pre-trained parametric \ac{PINN} can be used to calibrate constitutive models in both a deterministic and statistical setting. Note that in the synthetic test cases, the data for calibration is also generated using \ac{FEM}.}
    \label{fig:workflow}
\end{figure}
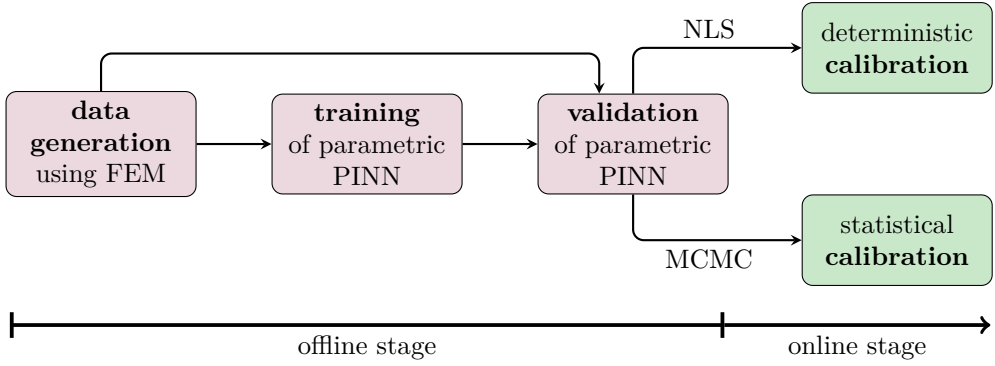

In this section, we describe the two test cases in more detail, specify the hyperparameters of the parametric \acp{PINN}' architecture and the training settings, and report the accuracy of the parametric forward solutions. In both test cases, we consider a plate with a hole. Since the geometry is two-fold symmetric, we consider only the top left quadrant of the plate and define symmetry boundary conditions on the bottom and right boundaries. We load the plate on the left edge with $\barvec{t} = [\SI{-100}{\N\per\mm\squared}, \SI{0}{\nounit}]^{\top}$.  Furthermore, external specific body forces, such as  gravity, are neglected. The geometry and boundary conditions are shown in \cref{fig:geometry_plate_with_hole}. The general workflow including data generation, training and validation of the parametric \ac{PINN} as well as calibration is outlined in \cref{fig:workflow} and explained in more detail in the following.

\subsubsection{Test case 1: Linear elasticity}\label{subsec:testcase_synthetic_linearelasticity}
As our first synthetic test case, we assume isotropic, linear elastic material and take construction steel as an example. Typical bulk and shear moduli for construction steel are $K=\SI{175000}{\N\per\mm\squared}$ and $G=\SI{80769}{\N\per\mm\squared}$, respectively, corresponding to a Young's modulus $E=\SI{210000}{\N\per\mm\squared}$ and Poisson's ratio $\nu=\SI{0.3}{\nounit}$, respectively. The plate is assumed to be under plane stress condition. \\

\begin{figure}[htb]
    \centering
    \begin{tikzpicture}[scale=0.8]
        \newcommand\XO{0}
        \newcommand\YO{0}
        \coordinate (A) at (\XO,\YO);
        \coordinate (B) at (\XO+4.5,\YO);
        \coordinate (C) at (\XO+5,\YO+0.5);
        \coordinate (D) at (\XO+5,\YO+5);
        \coordinate (E) at (\XO,\YO+5);
        \draw (A) -- (B);
        \draw (C) arc (90:180:0.5);
        \draw (C) -- (D);
        \draw (D) -- (E);
        \draw (E) -- (A);
        \draw[->, thick,color=red!70] (\XO,\YO+0) -- ++(180:1);
        \draw[->, thick,color=red!70] (\XO,\YO+1) -- ++(180:1);
        \draw[->, thick,color=red!70] (\XO,\YO+2) -- ++(180:1);
        \draw[->, thick,color=red!70] (\XO,\YO+3) -- ++(180:1);
        \draw[->, thick,color=red!70] (\XO,\YO+4) -- ++(180:1);
        \draw[->, thick,color=red!70] (\XO,\YO+5) -- ++(180:1);
        \draw[thick,color=red!70] (\XO-1,\YO) -- (\XO-1,\YO+5);
        \node (BC_bottom)[below] at (\XO+2.5,\YO-0.75) {$\barsca{u}_{y}=\SI{0}{\nounit}, \barsca{P}_{xy}=\SI{0}{\nounit}$};
        \node (BC_right_u)[right] at (\XO+5.75,\YO+2.25) {$\barsca{u}_{x}=\SI{0}{\nounit}$};
        \node (BC_right_t)[right] at (\XO+5.75,\YO+2.75) {$\barsca{P}_{yx}=\SI{0}{\nounit}$};
        \node (BC_left)[left] at (\XO-1.25,\YO+2.5) {$\barvec{t}=\begin{bmatrix}
            \SI{-100}{\N\per\mm\squared} \\
            \SI{0}{\nounit}
        \end{bmatrix}$};
        \node (BC_top)[above] at (\XO+2.5,\YO+5.25) {$\barvec{t}=\bm{0}$};
        \node (BC_hole)[below right] at (\XO+5.0,\YO) {$\barvec{t}=\bm{0}$};
        \draw[<->] (\XO,\YO-0.6) -- (\XO+5,\YO-0.6) node[midway, above, ] {$L = \SI{100}{\mm}$};
        \draw[<->] (\XO+5.6,\YO) -- (\XO+5.6,\YO+5) node[midway, above, rotate=90 ] {$L = \SI{100}{\mm}$};
        \draw[->] (\XO+5,\YO) -- ++(135:0.5) node[left] {R = \SI{10}{\mm}};
        \draw[->] (2.5,2.5) -- ++(90:1) node[midway, left] {y};
        \draw[->] (2.5,2.5) -- ++(0:1) node[midway, below] {x};
    \end{tikzpicture} 
    \caption{Geometry and boundary conditions of the top left quadrant of a plate with a hole under uniaxial tension. Body forces are neglected.}
    \label{fig:geometry_plate_with_hole}
\end{figure}
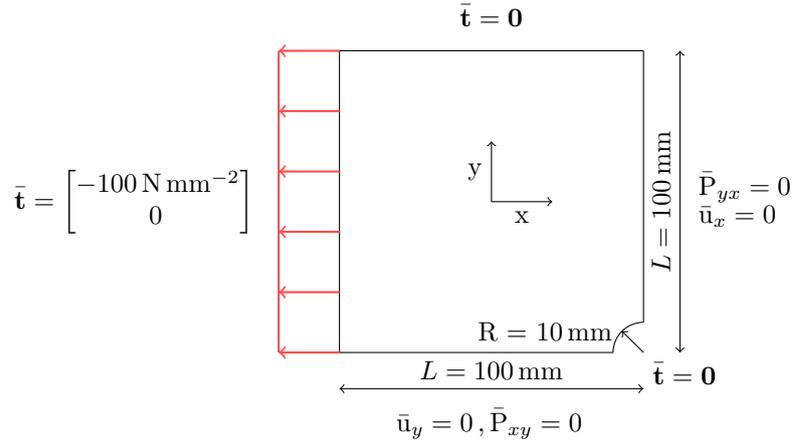

\noindent \textbf{FE simulations:} The synthetic displacement data for the training, validation and calibration data sets are generated by \ac{FE} simulations. For the \ac{FE} simulations, the geometry is meshed with triangular elements and we choose linear ansatz functions with one point integration. The displacement field is calculated and recorded at a total of $\SI{1148975}{\nounit}$ nodes. Due to the high resolution of the computational grid, the discretization errors are considered negligible.\\

\noindent \textbf{\ac{PINN}'s architecture and training:} We use a fully-connected \ac{FFNN} with six hidden layers each with $\SI{128}{\nounit}$ neurons and a hyperbolic tangent activation function. The \ac{PINN} has further four input neurons for the $x$- and $y$-coordinate and the two material parameters which are the bulk and shear modulus. Correspondingly, the \ac{PINN} has two output neurons for the displacement in $x$- and $y$-direction. The weights and biases of the \ac{FFNN} are initialized according to Glorot normal initialization \cite{glorot_understandingDifficultyTrainingANNs_2010} and with zeros, respectively.

For solving the resulting optimization problem that arises during training, we choose the L-BFGS optimization algorithm \cite{liu_LBFGS_1989,broyden_BFGS_1970,fletcher_BFGS_1970,goldfarb_BFGS_1970,shanno_BFGS_1970}. The training data set is composed as follows: We train the parametric \ac{PINN} for bulk and shear moduli within the range $K_{\textrm{train}} = [\SI{100000}{\N\per\mm\squared}, \SI{200000}{\N\per\mm\squared}]$ and $G_{\textrm{train}} = [\SI{60000}{\N\per\mm\squared}, \SI{100000}{\N\per\mm\squared}]$ corresponding to ranges for Young's modulus and Poisson's ratio of $E_{\textrm{train}} = [\SI{150000}{\N\per\mm\squared}, \SI{257143}{\N\per\mm\squared}]$ and $\nu_{\textrm{train}} = [\SI{0.125}{\nounit}, \SI{0.3636}{\nounit}]$, respectively. Therefore, we collect collocation points within the domain and on the boundaries for $\SI{1024}{\nounit}$ different combinations of bulk and shear moduli. These material parameter samples are obtained from the Sobol sequence \cite{sobol_sobolSampling_1967} from the material parameter domain. For each of the parameter samples, we generate $\SI{64}{\nounit}$ collocation points to enforce the \ac{PDE} ($\dataSetC$) within the domain and $\SI{64}{\nounit}$ collocation points on each of the five boundary segments ($\dataSetN$). While the collocation points on the boundaries are distributed uniformly, the collocation points within the domain are again obtained from a Sobol sequence. The stress boundary conditions are enforced as defined in \cref{fig:geometry_plate_with_hole}. Since we consider the strong form of the \ac{PDE}, it is essential to explicitly account for the symmetry stress boundary conditions on the bottom and right boundaries. Note that these symmetry boundary conditions are also imposed in the Galerkin \ac{FEM}. For the derivation of the correct boundary conditions, please refer to \cref{sec:appendix_bcs}. We further enhance the training by pre-simulated \ac{FE} data ($\dataSetd$) for $\SI{128}{\nounit}$ parameter samples obtained from the Sobol sequence from the parameter domain. For each of the parameter samples, we randomly pick $\SI{128}{\nounit}$ nodes from the \ac{FE} solution. In order to account for the different scales of the loss terms, we weight the data loss term by a constant factor $\lambdad = 10^{4}$. The runtime of training the \ac{PINN} on a NVIDIA \ac{GPU} A100 with 80 GB memory is about 14 hours.\\

\noindent \textbf{Validation:} For the validation of the parametric \ac{PINN} and the subsequent calibration, we generate a total of $\SI{100}{\nounit}$ different synthetic displacement data sets for randomly selected combinations of bulk and shear moduli using \ac{FE} simulations. We do not expect the parametric \acp{PINN} to approximate the displacements well beyond the training range of the material parameters. To prevent the realizations of the material parameters from being too close to the edges of the training range in calibration, we use a slightly limited parameter range for the generation of the synthetic full-field data. For the linear elastic constitutive model, we select bulk and shear moduli within the ranges $K_{\textrm{valid}} = [\SI{101000}{\N\per\mm\squared}, \SI{199000}{\N\per\mm\squared}]$ and $G_{\textrm{valid}} = [\SI{60500}{\N\per\mm\squared}, \SI{99500}{\N\per\mm\squared}]$, respectively. The validation is then performed on $\SI{1024}{\nounit}$ points randomly selected from each of the \ac{FE} solutions. In comparison to the high-fidelity \ac{FE} solution, the \ac{MAE} and the \ac{rL2} of the parametric \ac{PINN} yield $\MAE = \SI{1.32 e-5}{\nounit}$ and $\rLTwo = \SI{9.98 e-4}{\nounit}$, respectively. Note that the calibration data is different from the data we use to enhance the training. Please refer to \cref{sec:appendix_error_measures} for a definition of the error measures used in our numerical tests.

\subsubsection{Test case 2: Hyperelasticity}\label{subsec:testcase_synthetic_hyperelasticity}
In the second synthetic test case, we assume a weakly compressible Neo-Hookean material. The geometry of the plate with a hole and the boundary conditions are the same as in test case 1, see \cref{fig:geometry_plate_with_hole}. We assume the plate to be under plane strain condition.\\

\noindent \textbf{\ac{FE} simulations:} For the generation of the \ac{FE} data, we mesh the geometry with triangular elements, but choose quadratic ansatz functions with four quadrature points. The \ac{FE} solution is computed and recorded at a total of $\SI{1150118}{\nounit}$ nodes and we consider discretization errors to be negligible. \\

\noindent \textbf{\ac{PINN}'s architecture and training:} The hyperparameters of the parametric \ac{PINN} and the training settings as well as the number and composition of the training and validation data sets are defined identically to test case 1 except for the training ranges of the material parameters. For the hyperelastic material, we consider bulk and shear moduli within the range $K_{\textrm{train}} = [\SI{4000}{\N\per\mm\squared}, \SI{8000}{\N\per\mm\squared}]$ and $G_{\textrm{train}} = [\SI{500}{\N\per\mm\squared}, \SI{1500}{\N\per\mm\squared}]$. The runtime required for training the \ac{PINN} on a NVIDIA \ac{GPU} A100 with 80 GB memory is about 45 hours. Furthermore, it should be noted that a non-physical behavior due to the compressible part of the strain-energy function \cref{eq:strain_energy_volumetric} is not observable in the chosen parameter range. For details, see \cite{ehlers_largeVolumetricStrainsHyperelasticity_1998,hartmann_generalizedStrainEnergyFunctions_2003}.\\

\noindent \textbf{Validation:} As in test case 1, we generate a total of $\SI{100}{\nounit}$ different synthetic displacement data sets using \ac{FE} simulations. In parameter space, we randomly sample bulk and shear moduli within the ranges $K_{\textrm{valid}} = [\SI{4020}{\N\per\mm\squared}, \SI{7980}{\N\per\mm\squared}]$ and $G_{\textrm{valid}} = [\SI{505}{\N\per\mm\squared}, \SI{1495}{\N\per\mm\squared}]$, respectively. For validation, we use $\SI{1024}{\nounit}$ points randomly selected from each of the \ac{FE} solutions. In relation to the validation data, the parametric \ac{PINN} yields a \ac{MAE} and a \ac{rL2} of $\MAE = \SI{4.92 e-5}{\nounit}$ and $\rLTwo = \SI{1.04 e-4}{\nounit}$, respectively.

\subsection{Deterministic calibration}\label{subsec:results_synthetic_deterministic}
In the following, we present the results for the deterministic \ac{NLS} calibration for the two synthetic test cases. For the formulation of the \ac{NLS} calibration problem, please refer to \cref{subsec:problem_statement_deterministic}. In order to make robust statements about the accuracy of deterministic calibration, the accuracy of the identified material parameters for a total of $\SI{100}{\nounit}$ synthetic full-field displacement measurements is statistically analyzed. For the deterministic calibration, we use the L-BFGS algorithm and initialize the material parameters with the mean value of their training range, respectively.

We test the calibration for the same synthetic data sets that we used to validate the performance of the parametric \ac{PINN}, see \cref{subsec:testcases_synthetic}. In contrast to validation, however, we add artificial noise. First, we select $\SI{128}{\nounit}$ data points at random from each of the $\SI{100}{\nounit}$ synthetic full-field measurements. Second, in order to emulate real \ac{DIC} data, we add Gaussian noise $\mathcal{N}(0,\sigma^{2})$ with zero mean to the clean synthetic displacement data. According to \cite{pierron_extensionVirtualFieldsMethod_2010,hartmann_temperatureGradientDetermination_2023}, the noise in \ac{DIC} images has a standard deviation of $\sigma=\SI{4e-4}{\mm}$. To take into account that the optimal conditions required for this value are not always achieved in practice, we assume a standard deviation of $\sigma=\SI{5e-4}{\mm}$ instead.

In \cref{tab:results_deterministic_synthetic}, the results for test cases 1 and 2 are listed. We report the mean \acp{ARE} of the identified parameters compared to the true parameters used to calculate the synthetic data. In addition, to be able to estimate the scatter of the results, we also provide the \acp{SEMs} as well as the minimum and maximum \acp{ARE}. For a definition of the error measures used to evaluate the calibration results, please see \cref{sec:appendix_error_measures}. 

The results show that for both the linear elastic and the hyperelastic constitutive model, the material parameters can be identified with only small \acp{ARE}. In addition, the scatter of the \acp{ARE} is small in both test cases, as evidenced by the \acp{SEMs}. However, for the hyperelastic constitutive model, the errors are even significantly smaller than for the linear elastic constitutive model. We suspect that one reason for this observation is different ratios between the magnitude of the noise and the absolute displacements in the two test cases. The order of magnitude of the maximum absolute displacements in both $x$- and $y$- direction is $\mathcal{O}(10^{-2})$ in test case 1 (linear elasticity) and $\mathcal{O}(10^{0})$ in test case 2 (hyperelasticity) and is thus two orders of magnitude higher. At the same time, the magnitude and standard deviation of the noise remains constant, as these are only associated with the device, not with the observations. Hence, in test case 1, the noise has a significantly greater influence. Another reason for the larger \acp{ARE} and \acsp{SEM} for calibrating the linear elastic constitutive model is that the parametric \ac{PINN} in test case 1 is trained for a significantly larger parameter range. For both test cases, the \ac{NLS} calibration takes less than five seconds on average on a NVIDIA \ac{GPU} A100 with 80 GB memory. The number of parametric \ac{PINN} evaluations per calibration is $\mathcal{O}(10^{1})$ in both test cases.

In addition, we compare the calibration with parametric \acp{PINN} against the calibration with lookup tables where we estimate the material parameters from the \ac{FE} training data. We always compare against the best \ac{FE} data point, i.e., the one with the smallest Euclidean distance to the true material parameters in the parameter space. Under this assumption, the \acp{MARE} using the lookup table for test case 1 (linear elasticity) are $\MARE = \SI{1.02}{\percent}$ and $\MARE = \SI{1.98}{\percent}$ for the bulk and shear modulus, respectively. In test case 2 (hyperelasticity), the difference in accuracy is even more significant. Estimation of bulk and shear moduli using the lookup table results in $\MARE = \SI{0.88}{\percent}$ and $\MARE = \SI{5.07}{\percent}$, respectively. The comparison clearly shows that the parametric \ac{PINN} can especially approximate the nonlinear material behavior better than a simple lookup table that contains the \ac{FE} training data.

\begin{table}[ht]
\centering
\caption{Results of deterministic \ac{NLS} calibration for the synthetic displacement data in test cases 1 and 2. We repeat the \ac{NLS} calibration for $\SI{100}{\nounit}$ synthetic \ac{DIC} measurements for different combinations of material parameters. From the obtained $\SI{100}{\nounit}$ identified material parameter sets, we calculate the mean \acfp{ARE} with respect to the exact material parameters used for data generation. In addition, we provide the \acfp{SEMs} as well as the minimum and maximum \acp{ARE} to be able to estimate the scatter of the errors.}
\begin{tabular}{p{20mm}p{22mm}C{16mm}C{16mm}C{16mm}C{16mm}} \toprule
                                                            &                   & \multicolumn{4}{c}{\acf{ARE} [\%]} \\ \cmidrule{3-6}
                                                            &                       & mean                          & \acs{SEM}                     & minimum                       & maximum   \\ \midrule
                            
    \multirow{2}{22mm}{test case 1: linear elasticity}      & bulk modulus $K$      & $\SI{7.20 e-1}{\nounit}$      & $\SI{5.41 e-2}{\nounit}$      & $\SI{1.09 e-2}{\nounit}$      & $\SI{2.63}{\nounit}$       \\
                                                            & shear modulus $G$     & $\SI{1.57 e-1}{\nounit}$      & $\SI{1.18 e-2}{\nounit}$      & $\SI{6.86 e-4}{\nounit}$      & $\SI{4.79 e-1}{\nounit}$   \\ \midrule
    \multirow{2}{22mm}{test case 2: hyperelasticity}        & bulk modulus $K$      & $\SI{1.23 e-2}{\nounit}$      & $\SI{1.03 e-3}{\nounit}$      & $\SI{1.23 e-5}{\nounit}$      & $\SI{5.83 e-2}{\nounit}$   \\
                                                            & shear modulus $G$     & $\SI{1.64 e-3}{\nounit}$      & $\SI{1.27 e-4}{\nounit}$      & $\SI{7.47 e-8}{\nounit}$      & $\SI{5.68 e-3}{\nounit}$       \\ \bottomrule
\end{tabular}
\label{tab:results_deterministic_synthetic}
\end{table}

\subsection{Bayesian statistical inference}\label{subsec:results_synthetic_statistical}

In this subsection, we address the model calibration problem from a Bayesian statistical point of view. We treat the material parameters as random variables with a prior distribution that represents our estimate of the material parameters before we have seen the data. We then perform Bayesian statistical inference and sample the posterior distribution performing a \ac{MCMC} analysis. In order to validate the uncertainty of the estimated parameters from a frequentist point of view, we further carry out a coverage test. For the detailed formulation of the statistical calibration problem, we refer to \cref{subsec:problem_statement_statistical}.

We carry out a coverage test for a total of $\SI{100}{\nounit}$ synthetic full-field displacement measurements to validate the $\SI{95}{\percent}$-credible interval of the sampled posterior distributions. We use the same synthetic data as in the deterministic calibration. To emulate real \ac{DIC} data, we add Gaussian noise $\mathcal{N}(0,\sigma^{2})$ with zero mean and standard deviation $\sigma=\SI{5e-4}{\mm}$ to the clean synthetic displacement data. As we lack more detailed prior knowledge, we employ uniform priors covering the parameter range in which the parametric \acp{PINN} were trained. The \ac{MCMC} analysis is performed using the \texttt{emcee} algorithm. For both test cases, we employ an ensemble of $\SI{100}{\nounit}$ workers each with a chain length of $\SI{200}{\nounit}$. The workers are initialized randomly within the material parameter training ranges. Before the parameter samples are recorded, we run a burn-in phase with a chain length of $\SI{100}{\nounit}$ for each worker. In the burn-in phase, the Markov chain explores the parameter space and the drawn samples are not representative for the posterior distribution. We further choose a stretch scale of $\SI{4}{\nounit}$ which results in sound acceptance ratios that should be between $\SI{0.2}{\nounit}$ and $\SI{0.5}{\nounit}$ as a rule of thumb \cite{foremanMackey_emcee_2013}. 

\begin{figure}[ht]
    \newcommand{\subfigureWidth}{0.49\linewidth}
    \centering
    \begin{subfigure}{\subfigureWidth}
        \centering
        \includegraphics[width=1\linewidth]{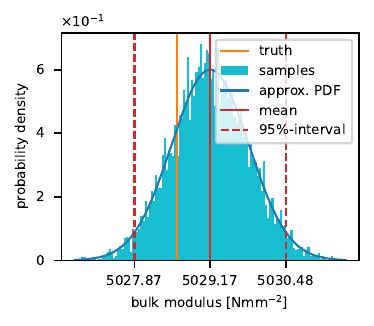}
        \caption{}
    \end{subfigure}
    \begin{subfigure}{\subfigureWidth}
        \centering
        \includegraphics[width=1\linewidth]{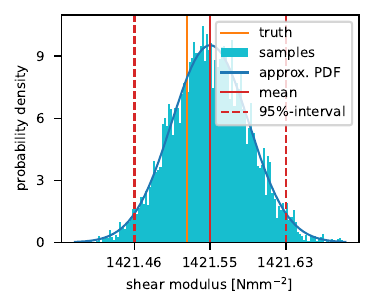}
        \caption{}
    \end{subfigure}
    
    \caption{Exemplary histograms of the posterior distribution of (a) bulk and (b) shear modulus for the hyperelastic constitutive model determined by Bayesian statistical inference. The illustration shows exemplary that the assumption of normally distributed posteriors is reasonable.}
    \label{fig:histograms_synthetic}
\end{figure}

The results of the Bayesian statistical inference are listed in \cref{tab:results_synthetic_statistical}. The coverage test clearly shows that the estimated uncertainty is valid in the sense of frequentist statistics. For both test cases 1 and 2, the coverage for both material parameters is close to the expected $\SI{95}{\percent}$. We further report the average bias of the posterior mean values with respect to the true material parameters and the standard deviations of the posterior distributions. To calculate the coverage, we have made the assumption that the sampled posterior \ac{PDF} can be approximated by a Gaussian distribution. As shown in \cref{fig:histograms_synthetic} as an example, this is a reasonable assumption. Furthermore, the runtime for the \ac{MCMC} analysis is less than $\SI{60}{\nounit}$ seconds on average on a NVIDIA \ac{GPU} A100 with 80 GB memory. According to the hyperparameters of the \texttt{emcee} algorithm specified above, the parametric \ac{PINN} is evaluated a total of $\SI{3e4}{\nounit}$ times in each \ac{MCMC} analysis.

\begin{table}[ht]
\centering
\caption{
Results of Bayesian statistical inference for the synthetic displacement data in test cases 1 and 2. We carry out a coverage test comprising $\SI{100}{\nounit}$ synthetic \ac{DIC} measurements each for different combinations of material parameters. The coverage indicates the percentage of test cases for which the true material parameter used to generate the synthetic data is within the $\SI{95}{\percent}$-credible interval. We further report the average bias of the posterior mean values with respect to the true material parameters and the standard deviations of the posterior distributions.
}
\begin{tabular}{p{22mm}p{22mm}C{15mm}C{25mm}C{25mm}} \toprule
                                                            &                     & coverage                & average bias \newline of mean $[\SI{}{\N\per\mm\squared}]$    & standard deviation $[\SI{}{\N\per\mm\squared}]$                \\ \midrule
                            
    \multirow{2}{22mm}{test case 1: linear elasticity}      & bulk modulus $K$    & $\SI{94}{\percent}$     & $\SI{-147.65}{\nounit}$                                       & $\SI{1200.61}{\nounit}$   \\
                                                            & shear modulus $G$   & $\SI{92}{\percent}$     & $\SI{9.26}{\nounit}$                                          & $\SI{138.84}{\nounit}$    \\ \midrule
    \multirow{2}{22mm}{test case 2: hyperelasticity}        & bulk modulus $K$    & $\SI{93}{\percent}$     & $\SI{-2.26 e-1}{\nounit}$                                     & $\SI{9.10 e-1}{\nounit}$  \\
                                                            & shear modulus $G$   & $\SI{98}{\percent}$     & $\SI{2.73 e-3}{\nounit}$                                      & $\SI{2.44 e-2}{\nounit}$  \\ \bottomrule
\end{tabular}
\label{tab:results_synthetic_statistical}
\end{table}

\section{Results for experimental full-field data}\label{sec:results_experimental}

Finally, we showcase the calibration of the linear elastic constitutive model from real-world experimental full-field displacement data. As with the synthetic data in \cref{sec:results_synthetic}, we perform both a deterministic and a statistical calibration.

\subsection{Setup and training of parametric PINN}\label{subsec:testcase_experimental}
We consider experimental full-field displacement data measured in a tensile test using \ac{DIC}. In the experiment, we used a specimen of S235 steel and assume linear elastic material behaviour. \\

\begin{figure}[htb]
    \centering
    \begin{tikzpicture}[scale=1.0]
        \newcommand\XO{0}
        \newcommand\YO{0}
        \coordinate (A) at (\XO,\YO);
        \coordinate (B) at (\XO+2.5,\YO);
        \coordinate (C) at (\XO+3.25,\YO+0.25);
        \coordinate (D) at (\XO+7.75,\YO+0.25);
        \coordinate (E) at (\XO+8.5,\YO);
        \coordinate (F) at (\XO+11,\YO);
        \coordinate (G) at (\XO+11,\YO+1.5);
        \coordinate (H) at (\XO+8.5,\YO+1.5);
        \coordinate (I) at (\XO+7.75,\YO+1.25);
        \coordinate (J) at (\XO+3.25,\YO+1.25);
        \coordinate (K) at (\XO+2.5,\YO+1.5);
        \coordinate (L) at (\XO,\YO+1.5);
        \coordinate (M) at (\XO+5.5,\YO+0.75); 
        \coordinate (MA) at (\XO+3.5,\YO+0.25);
        \coordinate (MB) at (\XO+7.5,\YO+0.25);
        \coordinate (MC) at (\XO+7.5,\YO+1.25);
        \coordinate (MD) at (\XO+3.5,\YO+1.25);
        \newcommand\clampedareacolor{gray!20};
        \fill[\clampedareacolor] (A) rectangle (K);
        \fill[\clampedareacolor] (E) rectangle (G);
        \fill[red!40] (MA) rectangle (MC);
        \fill[white] (M) circle (0.2);
        \newcommand\geometrystyle{dotted};
        \draw[\geometrystyle] (A) -- (B);
        \draw[\geometrystyle] (C) arc (90:126.87:1.25);
        \draw[\geometrystyle] (C) -- (D);
        \draw[\geometrystyle] (D) arc (90:53.13:1.25);
        \draw[\geometrystyle] (E) -- (F);
        \draw[\geometrystyle] (F) -- (G);
        \draw[\geometrystyle] (G) -- (H);
        \draw[\geometrystyle] (I) arc (270:306.87:1.25);
        \draw[\geometrystyle] (I) -- (J);
        \draw[\geometrystyle] (J) arc (270:233.13:1.25);
        \draw[\geometrystyle] (K) -- (L);
        \draw[\geometrystyle] (L) -- (A);
        \draw[\geometrystyle] (M) circle (0.2);
        \newcommand\simplifiedgeometrycolor{black};
        \draw[\simplifiedgeometrycolor] (C) arc (90:126.87:1.25);
        \draw[\simplifiedgeometrycolor] (C) -- (MB);
        \draw[\simplifiedgeometrycolor] (MB) -- (MC);
        \draw[\simplifiedgeometrycolor] (MC) -- (J);
        \draw[\simplifiedgeometrycolor] (J) arc (270:233.13:1.25);
        \draw[\simplifiedgeometrycolor] (B) -- (K);
        \draw[\simplifiedgeometrycolor] (M) circle (0.2);
        \node (BC_left_ux)[left,color=\simplifiedgeometrycolor] at (\XO+2.5,\YO+1.0) {$\barsca{u}_{x}=\SI{0}{\nounit}$};
        \node (BC_left_uy)[left,color=\simplifiedgeometrycolor] at (\XO+2.5,\YO+0.5) {$\barsca{u}_{y}=\SI{0}{\nounit}$};
        \node (BC_right_t)[right,color=\simplifiedgeometrycolor] at (\XO+7.5,\YO+0.75) {$\barvec{t}=\begin{bmatrix}
            \SI{106.26}{\N\per\mm\squared} \\
            \SI{0}{\nounit}
        \end{bmatrix}$};
        \draw[<->] (\XO-0.25,\YO) -- (\XO-0.25,\YO+1.5) node[midway, below, rotate=270] {30 mm};
        \draw[<->] (\XO+11.25,\YO+0.25+0.01875) -- (\XO+11.25,\YO+1.25-0.01875) node[midway, above, rotate=270] {$\SI{20}{\mm}$};
        \draw[<->] (\XO,\YO+1.75) -- (\XO+2.5,\YO+1.75) node[midway, above] {\SI{50}{mm}};
        \draw[<->] (\XO+3.5,\YO-0.25) -- (\XO+7.5,\YO-0.25) node[midway, above] {$\SI{80}{mm}$};
        \draw[<->] (\XO+3.25,\YO+1.75) -- (\XO+7.75,\YO+1.75) node[midway, above] {\SI{90}{mm}};
        \draw[<->] (\XO,\YO-0.75) -- (\XO+11,\YO-0.75) node[midway, above] {\SI{220}{mm}};
        \draw[<-] (\XO+3.25-0.395,\YO+0.25-0.064) -- ++(288.43:0.5) node[left] {$\SI{25}{mm}$};
        \draw[->] (M) -- ++(135:0.2);
        \draw (\XO+5.7,\YO+0.75) node[right] {$\SI{4}{mm}$};     
    \end{tikzpicture} 
    \caption{Geometry and boundary conditions of the tensile test. The specimen is clamped on the left side and subjected to traction on the right side (the clamped areas are filled in gray). The displacements were measured by a \ac{DIC} system for the area filled in red. The parametric \ac{PINN} is trained for the boundary conditions shown in the figure and the simplified geometry defined by the solid lines. Free Neumann boundary conditions were applied at the upper and lower edge of the geometry and in the hole.} 
    \label{fig:setup_tensile_test}
\end{figure}
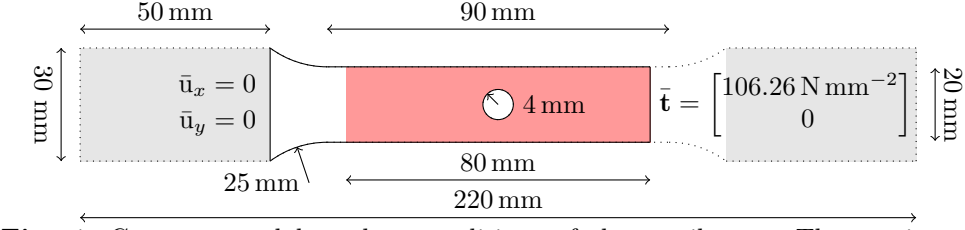

\noindent \textbf{Experimental settings:} The specimen was clamped on the left side and the testing machine pulled on the right side in axial direction up to an averaged axial strain of $\meanSup{\varepsilon} = \SI{5.1 e-2}{\percent}$. Thus, the strain is still in the linear elastic regime of the material under consideration. After a maximum traction of $\barvec{t} = [\SI{106.26}{\N\per\mm\squared}, 0 ]^{\top}$ had been applied, the displacements in the parallel area around the hole were measured with a \ac{DIC} system. For an illustration of the specimen geometry, the boundary conditions and the measurement area, please refer to \cref{fig:setup_tensile_test}. The full-field \ac{DIC} measurement is published in \cite{troeger_DICMeasurementLinearElasticSteel_2024}. \\

\noindent \textbf{\ac{FE} simulations:} To enhance the training process and to validate the parametric \ac{PINN}, we generate high fidelity displacement data using \ac{FEM}. Therefore, the simplified geometry is meshed with triangular elements and we choose linear ansatz functions with one point integration. The displacement field is then calculated and recorded for a total of $\SI{232984}{\nounit}$ nodes. Discretization errors are neglected due to the high resolution of the computational grid. \\

\noindent \textbf{\ac{PINN}'s architecture and training:} The hyperparameters of the parametric \ac{PINN} and the training settings are identical to the two previous test cases. To reduce the complexity, we train the parametric \ac{PINN} not for the complete specimen geometry but for a simplified one, see \cref{fig:setup_tensile_test}. For this purpose, we transfer the stress boundary condition from the end of the clamped area where the traction was actually applied to the end of the parallel area. As a prerequisite, we make the assumption that the force is distributed homogeneously over the height of the sample. The training of the \ac{PINN} on a NVIDIA \ac{GPU} A100 with 80 GB memory takes about 32 hours.

The training data is composed as follows: We train the parametric \ac{PINN} for bulk and shear moduli in the range $K_{\textrm{train}} = [\SI{100000}{\N\per\mm\squared}, \SI{200000}{\N\per\mm\squared}]$ and $G_{\textrm{train}} = [\SI{60000}{\N\per\mm\squared}, \SI{100000}{\N\per\mm\squared}]$ corresponding to ranges for Young's modulus and Poisson's ratio of $E_{\textrm{train}} = [\SI{150000}{\N\per\mm\squared}, \SI{257143}{\N\per\mm\squared}]$ and $\nu_{\textrm{train}} = [\SI{0.125}{\nounit}, \SI{0.3636}{\nounit}]$, respectively. For training, we consider $\SI{1024}{\nounit}$ different combinations of the material parameters obtained from a Sobol sequence. For each of the parameter samples, we generate $\SI{64}{\nounit}$ collocation points within the domain ($\dataSetC$) and $\SI{64}{\nounit}$ collocation points on each of the six boundary segments ($\dataSetN$). In addition, we enhance the training data set by pre-simulated \ac{FE} data ($\dataSetd$). We randomly select $\SI{128}{\nounit}$ data points from the \ac{FEM} solution each for $\SI{128}{\nounit}$ material parameter combinations obtained from a Sobol sequence. We further weight the data loss term by $\lambdad = 10^{6}$ in order to account for the different loss term scales. \\

\noindent\textbf{Validation:} As in the previous test cases, validation is performed on $\SI{1024}{\nounit}$ data points directly and randomly taken from the \ac{FEM} solution each for $\SI{100}{\nounit}$ randomly sampled parameter combinations within the training ranges. In relation to the validation data, the parametric \ac{PINN} yields a \ac{MAE} and a \ac{rL2} of $\MAE = \SI{1.08 e-6}{\nounit}$ and $\rLTwo = \SI{6.32 e-5}{\nounit}$, respectively.

\subsection{Deterministic calibration}\label{subsec:results_experimental_deterministic}
The full-field displacement measurement comprises a total of $\SI{5244}{\nounit}$ data points within the parallel area around the hole, see \cref{fig:setup_tensile_test} for the specimen geometry. For calibration, we again use the L-BFGS algorithm and initialize the material parameters with the mean value of their training range, respectively. As reference solution, we use the result of a \ac{NLS-FEM} calibration. In this approach, the parameters-to-state map is realized by a \ac{FE} simulation that is performed in each iteration instead of using the parametric \ac{PINN}. For solving the \ac{NLS-FEM} problem, the \texttt{lsqnonlin} function in Matlab is used. For more information on this approach when using full-field displacement data, please refer to \cite{roemer_modelCalibrationInSolidMechanics_2024}.

\begin{figure}[htb]
    \newcommand{\subfigureWidth}{0.95\linewidth}
    \centering
    \begin{subfigure}{\subfigureWidth}
        \centering
        \includegraphics[width=1\linewidth]{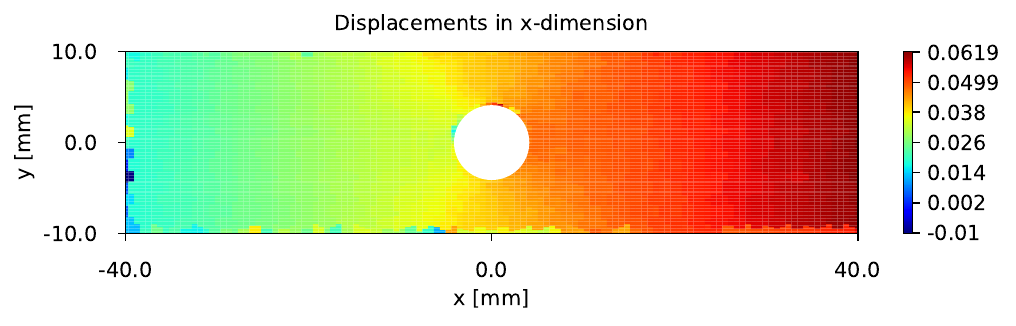}
        \caption{}
    \end{subfigure}
    \begin{subfigure}{\subfigureWidth}
        \centering
        \includegraphics[width=1\linewidth]{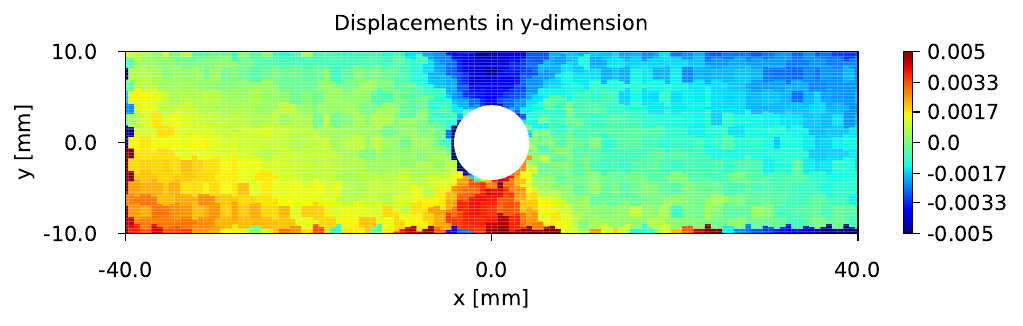}
        \caption{}
    \end{subfigure}
    \caption{Visualization of the displacements in (a) $x$-direction and (b) $y$-direction measured in the tensile test by \ac{DIC}. For visualization purposes, the measured displacements are interpolated onto a regular grid. Since the outliers significantly distort the scale of the displacements in $y$-direction, we limit the scale of the displacements in $y$-direction to $\sca{u}_{y}^{\textrm{visual}} = [\SI{-5 e-3}{\mm}, \SI{5 e-3}{mm}]$ for visualization purposes only.}
    \label{fig:dic_measurements_experimental}
\end{figure}

For the visualization of the \ac{DIC} images in \cref{fig:dic_measurements_experimental}, the measured displacements are interpolated onto a regular grid. The visualization shows that particularly in the area around the hole and the clamping, displacements were measured that deviate significantly from the expected displacement field. Since the outliers also significantly distort the scale of the displacements in $y$-direction, we therefore limit the scale of the displacements in $y$-direction to $\sca{u}_{y}^{\textrm{visual}} = [\SI{-5 e-3}{\mm}, \SI{5 e-3}{mm}]$ for visualization purposes only. In addition, it becomes clear that the measured displacements in $y$-direction are superimposed by a lateral displacement which may result from an eccentric clamping of the test specimen. However, it should be noted that the expected magnitude of the displacements in $y$-direction is small compared to the $x$-direction due to the material properties and the experimental setup. The measurement in $y$-direction is therefore more susceptible to external disturbances.

The results of the \ac{NLS} calibration are listed in \cref{tab:results_deterministic_experimental}. The calibration using the raw \ac{DIC} data yields a bulk and shear modulus of $K = \SI{109343}{\N\per\mm\squared}$ and $G = \SI{71125}{\N\per\mm\squared}$, respectively. In relation to the \ac{NLS-FEM} results, the identified material parameters deviate by \acp{RD} of $\RD_{K} = \SI{-14.63}{\percent}$ and $\RD_{G} = \SI{-3.29}{\percent}$. We assume that the reason for the large deviation is that the displacement data is pre-processed in \ac{NLS-FEM}. The measured full-field displacement data is linearly interpolated onto the \ac{FE} mesh nodes. In this process, outliers in the full-field measurement are smoothed out. For the linear interpolation, the Matlab function \texttt{scatteredInterpolant} with default settings is used. 
The parametric \ac{PINN}, on the other hand, uses the raw measurement data without pre-processing. For a fair comparison, we therefore also carry out the calibration with the interpolated displacement measurements. After interpolation, the full-field displacement measurement comprises a total of $\SI{1124}{\nounit}$ data points. The calibration using the interpolated data results in a bulk and shear modulus of $K = \SI{126679}{\N\per\mm\squared}$ and $G = \SI{73444}{\N\per\mm\squared}$, respectively, which deviate by \acp{RD} of $\RD_{K} = \SI{-1.10}{\percent}$ and $\RD_{G} = \SI{-0.13}{\percent}$ from the \ac{NLS-FEM} results. Furthermore, with the parametric \ac{PINN}, the runtime for the \ac{NLS} calibration is less than five seconds on a NVIDIA \ac{GPU} A100 with 80 GB memory. Both the parametric \ac{PINN} and the \ac{FE} model are evaluated $\mathcal{O}(10^{1})$ times.

\begin{table}[ht]
\centering
\caption{Results of deterministic \ac{NLS} calibration for the experimental displacement data. In addition to the material parameters identified by the parametric \ac{PINN}, we also report the results of a \ac{NLS-FEM} calibration as a reference solution. The parametric \ac{PINN} is applied to both the raw full-field displacement data and the displacement data linearly interpolated to the \ac{FE} mesh nodes.}
\begin{tabular}{p{40mm}C{35mm}C{35mm}} \toprule
                                                                            & bulk modulus $K$                      & shear modulus $G$                 \\ \midrule
    FEM (interpolated data)                                                 & \SI{128085}{\N\per\mm\squared}        & \SI{73541}{\N\per\mm\squared}     \\ \midrule
    PINN (raw data)                                                         & \SI{109343}{\N\per\mm\squared}        & \SI{71125}{\N\per\mm\squared}     \\
    $(\PINNSup{\kappa_{i}} - \FEMSup{\kappa_{i}}) / \FEMSup{\kappa_{i}}$    & \SI{-14.63}{\percent}                 & \SI{-3.29}{\percent}             \\ \midrule
    PINN (interpolated data)                                                & \SI{126679}{\N\per\mm\squared}        & \SI{73444}{\N\per\mm\squared}     \\
    $(\PINNSup{\kappa_{i}} - \FEMSup{\kappa_{i}}) / \FEMSup{\kappa_{i}}$    & \SI{-1.10}{\percent}                  & \SI{-0.13}{\percent}             \\ \bottomrule
\end{tabular}
\label{tab:results_deterministic_experimental}
\end{table}

\subsection{Statistical calibration}\label{subsec:results_experimental_statistical}
Finally, we determine the posterior distribution of the material parameters in the linear elastic constitutive model for the real-world experimental full-field displacement data. A detailed description of the experimental setup is given in \cref{subsec:testcase_experimental}. In order to validate our results for the parametric \ac{PINN}, we compare the posterior distributions to the results with \ac{FEM} as parameters-to-state map. As we found out in \cref{subsec:results_experimental_deterministic}, for a fair comparison, we need to use the interpolated displacement data. Furthermore, for the \ac{MCMC} analysis, we employ the \texttt{emcee} algorithm with an ensemble of $\SI{100}{\nounit}$ workers each with a chain length of $\SI{200}{\nounit}$ and a stretch scale of $\SI{4}{\nounit}$. Samples are recorded after a burn-in phase with a chain length of $\SI{100}{\nounit}$ for each worker. The workers are initialized randomly within the material parameter training ranges.

In the first attempt, we assumed Gaussian noise $\mathcal{N}(0,\sigma^{2})$ with zero mean and standard deviation $\sigma=\SI{5e-4}{\mm}$ just like with the synthetic data. However, without further modifications, we have not obtained reasonable results for this noise level. We suspect two possible reasons for the failure of the \ac{MCMC} analysis: 
\begin{enumerate}[label=(\roman*)]
    \item First, the noise in the present data is superimposed by measurement artifacts, such as lateral displacements due to a possibly eccentric clamping of the specimen. Additionally, in \cref{fig:dic_measurements_experimental}, we can see some measurement outliers close to the boundary caused by errors in the facet-matching in consequence of a slightly incorrect placement of the tensile specimen with respect to the camera alignment. The resulting measurement error which is made up of the background noise and the measurement artifacts is therefore probably greater than the assumed value of $\sigma=\SI{5e-4}{\mm}$.
    \item Second, we assume that the noise levels for the present data are actually different in the $x$- and $y$- directions. One possible reason for this is the different resolution of the \ac{DIC} system in the different spatial directions. In addition, in the deterministic setting, we have already observed that weighting the residuals is essential for the calibration from experimental data.
\end{enumerate}

\noindent We therefore propose to use the diagonal covariance matrix obtained from relating the \ac{NLS} problem to the maximum a posteriori estimate, see \cref{eq:maximum_a_posteriori_estimate,eq:maximum_a_posteriori_estimate_likelihood_substituted}. If we use a uniform prior over the admissible set $\Omega_{\bm{\kappa}}$ of material parameters $\bm{\kappa}$, we restrict the statistical calibration problem to the same parameter set as the deterministic \ac{NLS} problem, see \cref{eq:nls_optimization_problem}. With a uniform prior, the logarithm of the prior $\log p(\bm{\kappa})$ in \cref{eq:maximum_a_posteriori_estimate_likelihood_substituted} is constant and can be neglected in the minimization problem. The maximum a posteriori estimate then simplifies to the so-called maximum likelihood estimate
\begin{equation}\label{eq:maximum_likelihood_estimate}
\begin{aligned}
    \bm{\kappa}^{*} = \argmax_{\bm{\kappa}} \likelihood(\bm{\kappa}) &= \argmin_{\bm{\kappa}} \big( -\log \likelihood(\bm{\kappa}) \big) \\
    &= \argmin_{\bm{\kappa}} \big( \frac{1}{2} \norm{ \hatUState(\bm{\kappa}) - \vec{d} }_{\bm{\Sigma}_{\vec{e}}^{-1}}^{2} \big).
\end{aligned}
\end{equation}
For uniform priors, the diagonal covariance matrix $\bm{\Sigma}_{\vec{e}}$ can then be related to the weight matrix $\vec{W}$ in the non-regularized \ac{NLS} problem \cref{eq:nls_loss} by
\begin{equation}\label{eq:relation_covariance_and_weights}
    \bm{\Sigma}_{\vec{e}} := 
    \begin{bmatrix}
        \bm{\Sigma}_{\vec{e}_{x}} & \vec{0} \\
        \vec{0} & \bm{\Sigma}_{\vec{e}_{y}}
    \end{bmatrix}
    = ( \vec{W}^{\top} \vec{W} )^{-1},
    \; \bm{\Sigma}_{\vec{e}} \elmm{2\numd}{2\numd},
\end{equation}
where the sub-covariance matrices $\bm{\Sigma}_{\vec{e}_{x}}, \bm{\Sigma}_{\vec{e}_{y}} \elmm{\numd}{\numd}$ for i.i.d. noise are defined as
\begin{equation}\label{eq:sub_covariance_submatrices}
    \bm{\Sigma}_{\vec{e}_{x}} = \sigma_{x}^{2} \ten{I} \;\; \text{and} \;\; \bm{\Sigma}_{\vec{e}_{y}} = \sigma_{y}^{2} \ten{I},
\end{equation}
with the identity matrix of size $\numd \times \numd$ and standard deviations $\sigma_{x}$ and $\sigma_{y}$ of Gaussian noise $\mathcal{N}(\vec{0}, \sigma_{x}^{2} \ten{I})$ and $\mathcal{N}(\vec{0}, \sigma_{y}^{2} \ten{I})$ in $x$- and $y$-direction, respectively.

In the following, we use a uniform prior for the material parameters to be inferred and derive the covariance matrix from \cref{eq:maximum_likelihood_estimate,eq:relation_covariance_and_weights,eq:sub_covariance_submatrices} as described above. For the weight matrix used in the \ac{NLS} problem, we finally obtain standard deviations $\sigma_{x} = \SI{0.0401}{\mm}$ and $\sigma_{y} = \SI{0.0017}{\mm}$ for i.i.d. Gaussian noise $\mathcal{N}(\vec{0}, \sigma_{x}^{2} \ten{I})$ and $\mathcal{N}(\vec{0}, \sigma_{y}^{2} \ten{I})$ in $x$- and $y$-direction, respectively.

\begin{figure}[ht]
    \newcommand{\subfigureWidth}{0.49\linewidth}
    \centering
    \begin{subfigure}{1\linewidth}
        \centering
        \begin{subfigure}{\subfigureWidth}
            \centering
            \includegraphics[width=\linewidth]{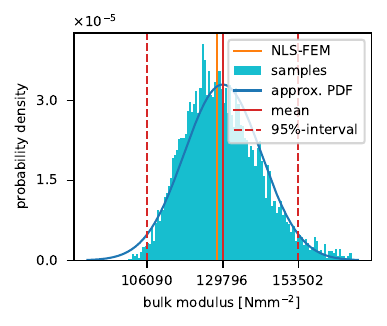}
        \end{subfigure}
        \begin{subfigure}{\subfigureWidth}
            \centering
            \includegraphics[width=\linewidth]{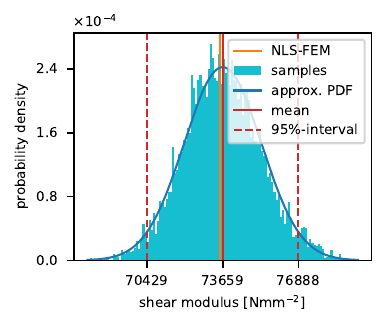}
        \end{subfigure}
        \caption{Parametric \ac{PINN}}
    \label{fig:histograms_experimental_pinn}
    \end{subfigure}

    \begin{subfigure}{1\linewidth}
        \centering
        \begin{subfigure}{\subfigureWidth}
            \centering
            \includegraphics[width=\linewidth]{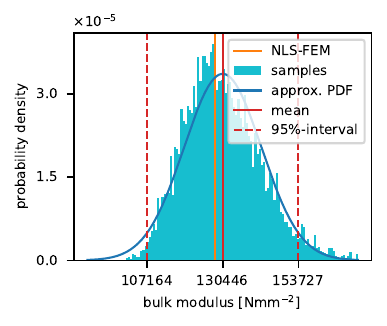}
        \end{subfigure}
        \begin{subfigure}{\subfigureWidth}
            \centering
            \includegraphics[width=\linewidth]{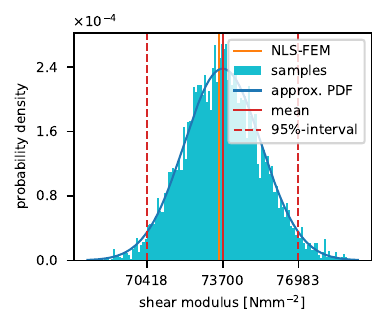}
        \end{subfigure}
        \caption{\ac{FEM}}
    \label{fig:histograms_experimental_fem}
    \end{subfigure}
    
    \caption{Posterior probability densities of bulk and shear modulus determined by a \ac{MCMC} analysis for the experimental displacement measurements. The results for the parametric \ac{PINN} in (a) show a good concentration of the probability density. For validation, in (b), we also provide the posterior probability densities we obtain when using \ac{FEM} as parameters-to-state map. The comparison shows a good level of agreement.}
    \label{fig:histograms_experimental}
\end{figure}

The posterior probability densities for bulk and shear modulus obtained by a \ac{MCMC} analysis are illustrated in \cref{fig:histograms_experimental_pinn}. The probability distributions show a good concentration and small uncertainties for both material parameters. Furthermore, the mean values of the posterior probability densities are close to the values we obtain from the deterministic \ac{NLS-FEM} calibration. This is expected since we derive the covariance matrix from the relation between the maximum a posteriori estimate and the \ac{NLS} problem. For validation, we also carry out the \ac{MCMC} analysis with \ac{FEM} as parameters-to-state map and the same covariance matrix, see \cref{fig:histograms_experimental_fem}. The comparison shows that the posterior probability densities obtained with the two different methods are in good agreement. Moreover, with the parametric \ac{PINN}, the runtime for the \ac{MCMC} analysis is less than $\SI{60}{\nounit}$ seconds on a NVIDIA \ac{GPU} A100 with 80 GB memory. According to the hyperparameters of the \texttt{emcee} algorithm specified above, the parametric \ac{PINN} is evaluated a total of $\SI{3e4}{\nounit}$ times in the \ac{MCMC} analysis.

In general, it is not straightforward to compare the runtimes of the parametric \acp{PINN} and \ac{NLS-FEM}, as we implemented both approaches in different programming languages and optimized them for specific hardware. However, to illustrate the difference in runtime, we perform the statistical calibration for the experimental data for both approaches on a \ac{CPU} in one thread. For this setting, we found that the statistical calibration using the pre-trained \ac{PINN} is approximately $\SI{300}{\nounit}$ times faster than using \ac{FE} as parameters-to-state map. For nonlinear problems, the speed-up is expected to be even more significant. Note that this hardware setting was only chosen for the sake of comparability. All other tests were carried out on the hardware for which the respective implementation was optimized.

Finally, we would like to make the following remarks: First, \acp{PINN} generally do not well extrapolate beyond the training domain. We therefore recommend the use of material parameter priors with at most weak support beyond the training range of the parametric \ac{PINN}. Otherwise, the Markov chain is more likely to explore regions in the parameter domain for which the parametric \ac{PINN} is not trained and thus does not provide good prediction accuracy. As mentioned before, in Bayesian inference, a correct uncertainty quantification relies on an accurate parameters-to-state map. Second, it should be noted that the noise levels derived from the weights used in the corresponding \ac{NLS} problem are not the actual noise levels of the measurements. The choice of the weights is usually based on heuristics and not necessarily on a statistical analysis of the measurement data. However, the chosen approach enables comparability between the statistical and the deterministic calibration problems. Third, we point out that Bayesian inference, in principle, also allows the noise level to be estimated simultaneously with the material parameters. Therefore, the noise can be modeled, e.g., by Gaussian distributions or by Gaussian processes \cite{rasmussen_gaussianProcesses_2005}. However, estimating the noise is beyond the scope of this work. For more information on this approach, we refer, for instance, to \cite{kennedy_bayesianCalibration_2001}.


\section{Conclusion and outlook}\label{sec:conclusion}

Advances in the development of full-field measurement capabilities, such as, e.g., \acf{DIC}, have recently led to an increasing interest in appropriate methods for the calibration of constitutive models. In experimental mechanics, the inverse problem of identifying the material parameters is traditionally solved by numerical methods, such as \acl{NLS-FEM} or \acl{VFM}. However, the computational costs associated with these methods are oftentimes to high, making them unsuitable for online applications. This results in an urgent need for methods that enable rapid calibration in online applications, even under severe time constraints.

In the present contribution, we demonstrate that the parametric \ac{PINN} approach enables an accurate and efficient model calibration and uncertainty quantification of the inferred material parameters. In the offline stage, the parametric \ac{PINN} is trained to learn a parameterized solution of the underlying parametric \acl{PDE} by encoding the physics into a loss function. In addition, training can be enhanced by high-fidelity simulation data that can be easily integrated into the training process. In the subsequent online stage, the parametric \ac{PINN} then can be employed as a surrogate for the parameters-to-state map in the calibration process. Due to the low computational costs of \acf{ANN} evaluations, calibration can be performed in near real-time, even though ten thousands of forward model evaluations are required. 

We demonstrated the advantages of using parametric \acp{PINN} for constitutive model calibration in deterministic \acl{NLS} calibration as well as \acf{MCMC}-based Bayesian inference in various numerical tests. First, we considered the calibration of a small strain linear elastic and a finite strain hyperelastic constitutive model using noisy synthetic data. A statistical evaluation of the results showed both high accuracy for the deterministic point estimate and valid uncertainty for the Bayesian inference. In addition, we calibrated a small strain linear elastic model using experimental full-field data from a tensile test. As reference, we used the results obtained when using the \acl{FEM} instead of the parametric \ac{PINN} as parameters-to-state map. The parametric \ac{PINN} also showed good results for the experimental data in both the deterministic and statistical settings. At the same time, the runtime of the parametric \ac{PINN} needed for online calibration is considerably shorter, especially when it comes to MCMC-based Bayesian inference.

To the best of the authors knowledge, this is the first contribution which presents parametric \acp{PINN} for the calibration of constitutive models. While it has often been stated that \acp{PINN} are especially suited for inverse problems, the settings considered in the literature so far are often far away from realistic applications. Herein, the authors have demonstrated the entire process from parametric \ac{PINN} training towards model calibration using real-world experimental data. The achieved savings in the online calibration step urge for further developments of parametric \acp{PINN} for more complex, history dependent and anisotropic materials. The pre-training of parametric \acp{PINN} may help to further establish full-field measurement techniques, such as \ac{DIC}, in materials development in both academia and industry as well as in online applications, such as continuous \acf{SHM}.

Although the parametric \acp{PINN} have already achieved good results in our numerical tests, further work is necessary for real-world applications. In the example with the experimental data, it became clear that the real measurement data can also contain measurement artifacts in addition to the background noise of the \ac{DIC} system. In contrast to the background noise, the measurement artifacts are difficult to characterize and make calibration more challenging. This applies in particular to \acp{PINN} as they usually use the data directly, without prior interpolation of the sensor data. For this reason, either a pre-processing of the data is necessary before calibration, or the additional uncertainties must be taken into account during calibration. Possible methods for pre-processing are, among others, polynomial interpolation \cite{rudy_dataDrivenDiscoveryOfPDEs_2017}, \ac{ANN}-based interpolation \cite{berg_dataDrivenDiscoveryOfPDEs_2019} or kernel methods \cite{flaschel_unsupervisedDiscoveryEUCLID_2021}. In a statistical setting, the measurement error could also be considered as an additional error term in \cref{eq:bayesian_reduced_observation} and modeled, e.g., by a Gaussian process \cite{kennedy_bayesianCalibration_2001}.

The authors are aware that a reliable measurement of full-field displacement data using, e.g., a \ac{DIC} system, places very high demands on the measurement system. These requirements are significantly higher for on-site online applications in \ac{SHM} compared to laboratory applications due to the environmental impacts acting on the system. However, the use of \ac{DIC} in the context of \ac{SHM} is an active field of research, see, e.g., \cite{dong_reviewVisionBasedSHM_2021, niezrecki_DICForSHM_2019,reagan_feasibilityOfDICInSHM_2018}.

From a modeling perspective, a further challenge arises as soon as the displacement or load boundary conditions are not constant. This is particularly likely for applications in the field of \ac{SHM}. The load boundary condition then needs to be inferred online using, e.g., load cells \cite{morenoGomez_sensorsUsedInSHM_2018}. However, every boundary condition that is not exactly known before training must be taken into account as a parameter and thus as an additional input to the parametric \ac{PINN}. This means that future work on methods for overcoming the curse of dimensionality are also of great importance.

\section*{Declarations}

\subsection*{Availability of data and materials}
The research code for both training of parametric \acp{PINN} and calibration is open-source and available both on GitHub and on Zenodo \cite{anton_codeParametricPINNsCalibration_2024}. The experimental dataset is available on Zenodo \cite{troeger_DICMeasurementLinearElasticSteel_2024}.


\subsection*{Competing interests}
The authors declare that they have no competing interests.


\subsection*{Funding}
DA, HW and UR acknowledge support by the Deutsche Forschungsgemeinschaft (DFG, German Research Foundation) in the project DFG 255042459: \textit{"GRK2075: Modeling the constitutional evolution of building materials and structures with respect to aging"}. DA and HW also acknowledge support in the project  DFG 501798687: \textit{"Monitoring data driven life cycle management with AR based on adaptive, AI-supported corrosion prediction for reinforced concrete structures under combined impacts"} which is a subproject of SPP 2388: \textit{"Hundred plus - Extending the Lifetime of Complex Engineering Structures through Intelligent Digitalization"} funded by the DFG. AH acknowledges support by an ETH Zurich Postdoctoral Fellowship.


\subsection*{Authors' contributions}
DA: conceptualization, data curation, formal analysis, investigation, methodology, project administration, software, validation, visualization, writing – original draft, writing – review and editing; JAT: data curation, investigation, software, validation, writing – original draft, writing – review and editing; HW: conceptualization, funding acquisition, methodology, resources, supervision, writing – original draft, writing – review and editing; UR: conceptualization, funding acquisition, methodology, supervision, writing – review and editing; AH: conceptualization, supervision, writing – review and editing; SH: resources, supervision, writing – review and editing.


\subsection*{Acknowledgements}
DA, HW and UR thank the members of the research training group \textit{GRK2075} for the fruitful discussions.


\section*{Acknowledgements}
This version of the article has been accepted for publication, after peer review but is not the Version of Record and does not reflect post-acceptance improvements, or any corrections. The Version of Record is available online at: https://doi.org/10.1186/s40323-025-00285-7


\appendix

\section{Boundary conditions in strong form PINNs}\label{sec:appendix_bcs}

We consider the balance equation \cref{eq:balance_momentum} with \acp{BC} \cref{eq:bcs} for the top left quadrant of a plate with a hole as described in \cref{subsec:testcases_synthetic}. The same test case has been considered earlier in \cite{li_PINNsForElasticPlates_2021}, where it has been reported that the accuracy of strong form \acp{PINN} was insufficient. Herein, we illustrate that the reason for the unsatisfactory results is merely an incomplete imposition of \acp{BC} in \cite{li_PINNsForElasticPlates_2021}. Note that in Galerkin finite element methods, Neumann \acp{BC} are treated via surface integrals, and zero traction \acp{BC} are automatically fulfilled. This is not the case for methods relying on the strong form.

\begin{figure}[htbp]
  \centering
  \begin{subfigure}[b]{1\linewidth}
    \centering
    \begin{tikzpicture}[scale=0.8]
        \newcommand\XO{0}
        \newcommand\YO{0}
        \coordinate (A) at (\XO,\YO);
        \coordinate (B) at (\XO+4.5,\YO);
        \coordinate (C) at (\XO+5,\YO+0.5);
        \coordinate (D) at (\XO+5,\YO+5);
        \coordinate (E) at (\XO,\YO+5);
        \draw (A) -- (B);
        \draw (C) arc (90:180:0.5);
        \draw (C) -- (D);
        \draw (D) -- (E);
        \draw (E) -- (A);
        \draw[->, thick,color=red!70] (\XO,\YO+0) -- ++(180:1);
        \draw[->, thick,color=red!70] (\XO,\YO+1) -- ++(180:1);
        \draw[->, thick,color=red!70] (\XO,\YO+2) -- ++(180:1);
        \draw[->, thick,color=red!70] (\XO,\YO+3) -- ++(180:1);
        \draw[->, thick,color=red!70] (\XO,\YO+4) -- ++(180:1);
        \draw[->, thick,color=red!70] (\XO,\YO+5) -- ++(180:1);
        \draw[thick,color=red!70] (\XO-1,\YO) -- (\XO-1,\YO+5);
        \node (BC_bottom)[below] at (\XO+2.5,\YO-0.75) {$\barsca{u}_{y}=\SI{0}{\nounit}$};
        \node (BC_right_u)[right] at (\XO+5.75,\YO+2.25) {$\barsca{u}_{x}=\SI{0}{\nounit}$};
        \node (BC_left)[left] at (\XO-1.25,\YO+2.5) {$\barvec{t}=\begin{bmatrix}
            \SI{-100}{\N\per\mm\squared} \\
            \SI{0}{\nounit}
        \end{bmatrix}$};
        \node (BC_top)[above] at (\XO+2.5,\YO+5.25) {$\barvec{t}=\bm{0}$};
        \node (BC_hole)[below right] at (\XO+5.0,\YO) {$\barvec{t}=\bm{0}$};
        \draw[<->] (\XO,\YO-0.6) -- (\XO+5,\YO-0.6) node[midway, above, ] {$L = \SI{100}{\mm}$};
        \draw[<->] (\XO+5.6,\YO) -- (\XO+5.6,\YO+5) node[midway, above, rotate=90 ] {$L = \SI{100}{\mm}$};
        \draw[->] (\XO+5,\YO) -- ++(135:0.5) node[left] {R = \SI{10}{\mm}};
        \draw[->] (2.5,2.5) -- ++(90:1) node[midway, left] {y};
        \draw[->] (2.5,2.5) -- ++(0:1) node[midway, below] {x};
    \end{tikzpicture} 
    \caption{\Acp{BC} as described in \cite{li_PINNsForElasticPlates_2021}. Only Dirichlet \acp{BC} are applied.}
    \label{fig:appendix_platewithhole_without_stress_symmetry_bcs}
  \end{subfigure}
  \hfill
  \begin{subfigure}[b]{1\linewidth}
    \centering
    \begin{tikzpicture}[scale=0.8]
        \newcommand\XO{0}
        \newcommand\YO{0}
        \coordinate (A) at (\XO,\YO);
        \coordinate (B) at (\XO+4.5,\YO);
        \coordinate (C) at (\XO+5,\YO+0.5);
        \coordinate (D) at (\XO+5,\YO+5);
        \coordinate (E) at (\XO,\YO+5);
        \draw (A) -- (B);
        \draw (C) arc (90:180:0.5);
        \draw (C) -- (D);
        \draw (D) -- (E);
        \draw (E) -- (A);
        \draw[->, thick,color=red!70] (\XO,\YO+0) -- ++(180:1);
        \draw[->, thick,color=red!70] (\XO,\YO+1) -- ++(180:1);
        \draw[->, thick,color=red!70] (\XO,\YO+2) -- ++(180:1);
        \draw[->, thick,color=red!70] (\XO,\YO+3) -- ++(180:1);
        \draw[->, thick,color=red!70] (\XO,\YO+4) -- ++(180:1);
        \draw[->, thick,color=red!70] (\XO,\YO+5) -- ++(180:1);
        \draw[thick,color=red!70] (\XO-1,\YO) -- (\XO-1,\YO+5);
        \node (BC_bottom)[below] at (\XO+2.5,\YO-0.75) {$\barsca{u}_{y}=\SI{0}{\nounit}, \barsca{P}_{xy}=\SI{0}{\nounit}$};
        \node (BC_right_u)[right] at (\XO+5.75,\YO+2.25) {$\barsca{u}_{x}=\SI{0}{\nounit}$};
        \node (BC_right_P)[right] at (\XO+5.75,\YO+2.75) {$\barsca{P}_{yx}=\SI{0}{\nounit}$};
        \node (BC_left)[left] at (\XO-1.25,\YO+2.5) {$\barvec{t}=\begin{bmatrix}
            \SI{-100}{\N\per\mm\squared} \\
            \SI{0}{\nounit}
        \end{bmatrix}$};
        \node (BC_top)[above] at (\XO+2.5,\YO+5.25) {$\barvec{t}=\bm{0}$};
        \node (BC_hole)[below right] at (\XO+5.0,\YO) {$\barvec{t}=\bm{0}$};
        \draw[<->] (\XO,\YO-0.6) -- (\XO+5,\YO-0.6) node[midway, above, ] {$L = \SI{100}{\mm}$};
        \draw[<->] (\XO+5.6,\YO) -- (\XO+5.6,\YO+5) node[midway, above, rotate=90 ] {$L = \SI{100}{\mm}$};
        \draw[->] (\XO+5,\YO) -- ++(135:0.5) node[left] {R = \SI{10}{\mm}};
        \draw[->] (2.5,2.5) -- ++(90:1) node[midway, left] {y};
        \draw[->] (2.5,2.5) -- ++(0:1) node[midway, below] {x};
    \end{tikzpicture}
    \caption{\Acp{BC} as described in \cite{henkes_PINNsForContinuumMechanics_2022}. Beside the Dirichlet \acp{BC}, the symmetry \acp{BC} also include Neumann \acp{BC} with respect to the shear stresses.}
    \label{fig:appendix_platewithhole_with_stress_symmetry_bcs}
  \end{subfigure}
  \caption{\Acp{BC} in the test case plate with a hole as described in (a) \cite{li_PINNsForElasticPlates_2021} and  (b) our formulation presented earlier in \cite{henkes_PINNsForContinuumMechanics_2022}.}
  \label{fig:appendix_platewithhole}
\end{figure}
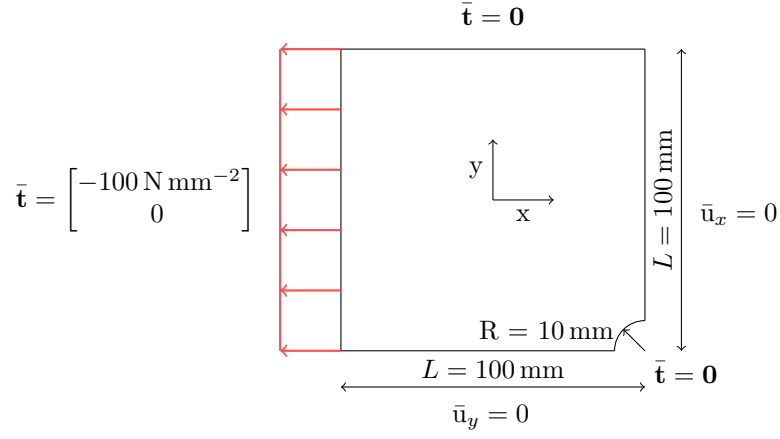
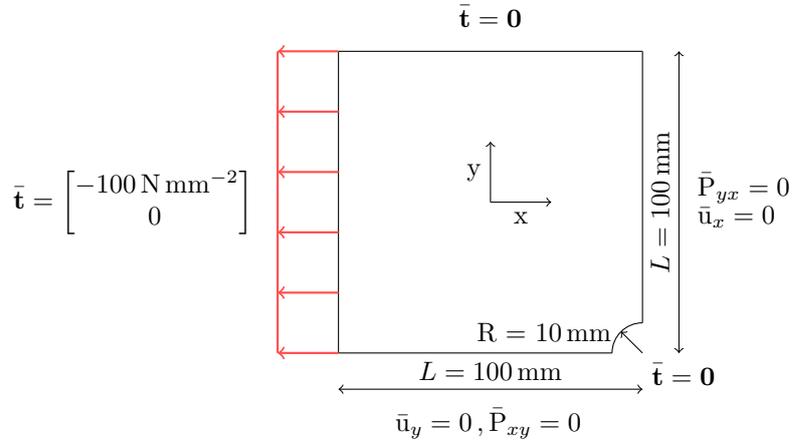

To this end, we exemplary consider the \textbf{right boundary} of the plate sketched in \cref{fig:appendix_platewithhole}, where the following \acp{BC} must be fulfilled:
\begin{subequations}
\begin{align}
    \sca{u}_x(x=0) &= 0, \label{eq:appendix_right_bc_dirichlet} \\
    \sca{P}_{yx}(x=0) &= 0. \label{eq:appendix_right_bc_neumann} 
\end{align}
\end{subequations}
In \cite{li_PINNsForElasticPlates_2021}, only the Dirichlet condition \cref{eq:appendix_right_bc_dirichlet} has been considered, see also \cref{fig:appendix_platewithhole_without_stress_symmetry_bcs}. However, since the balance of linear momentum \cref{eq:balance_momentum} results in two coupled \acp{PDE} for the considered 2D test case, at each  boundary two \acp{BC} need to be defined, one in each spatial dimension. With the surface normal of the right boundary $\vec{n}_{\textrm{right}} = [1, 0]^{\top}$, the Neumann \ac{BC} \cref{eq:appendix_right_bc_neumann} follows directly from $t_y = \SI{0}{\nounit}$:
\begin{equation}
\begin{aligned}\label{eq:appendix_neumann_right}
    \sca{t}_y = 0 &= \sca{P}_{yx} \sca{n}_x + \sca{P}_{yy} \sca{n}_y,    \\
          0 &= \sca{P}_{yx}.
\end{aligned}
\end{equation}

To illustrate that the correct application of \acp{BC} is essential, we solve the forward problem for the top left quadrant of a plate with a hole with and without symmetry stress \acp{BC} and compare the results. The geometry and \acp{BC} are shown in \cref{fig:appendix_platewithhole}. We use the ansatz \cref{eq:ppinn_ansatz} with a fully connected \acf{FFNN} with six hidden layers each with $\SI{64}{\nounit}$ neurons and hyperbolic tangent activation functions. The weights and biases of the \ac{FFNN} are initialized according to Glorot normal initialization \cite{glorot_understandingDifficultyTrainingANNs_2010} and with zeros, respectively. The training data set consists of $\SI{8192}{\nounit}$ collocation points within the domain and $\SI{256}{\nounit}$ points on each of the five boundary segments. No \ac{FE} data is used for training. We train the \ac{PINN} for a predefined bulk modulus $K=\SI{175000}{\N\per\mm\squared}$ and shear modulus $G=\SI{80769}{\N\per\mm\squared}$, respectively. The resulting optimization problem is solved using the L-BFGS optimization algorithm \cite{liu_LBFGS_1989,broyden_BFGS_1970,fletcher_BFGS_1970,goldfarb_BFGS_1970,shanno_BFGS_1970}.

\begin{figure}[htbp]
  \centering
  \begin{subfigure}[b]{0.49\linewidth}
    \includegraphics[width=\linewidth]{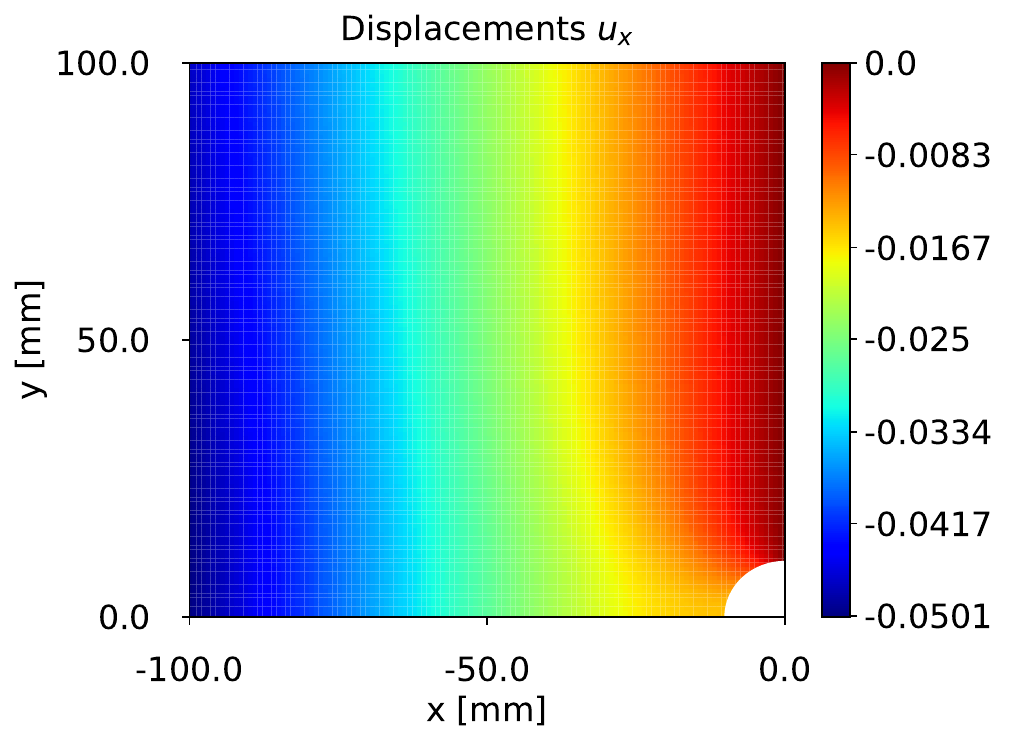}
    \caption{\ac{FEM} solution: Displacement field in $x$.}
    \label{fig:bc_comparison_fem_ux.pdf}
  \end{subfigure}
  \hfill
  \begin{subfigure}[b]{0.49\linewidth}
    \includegraphics[width=\linewidth]{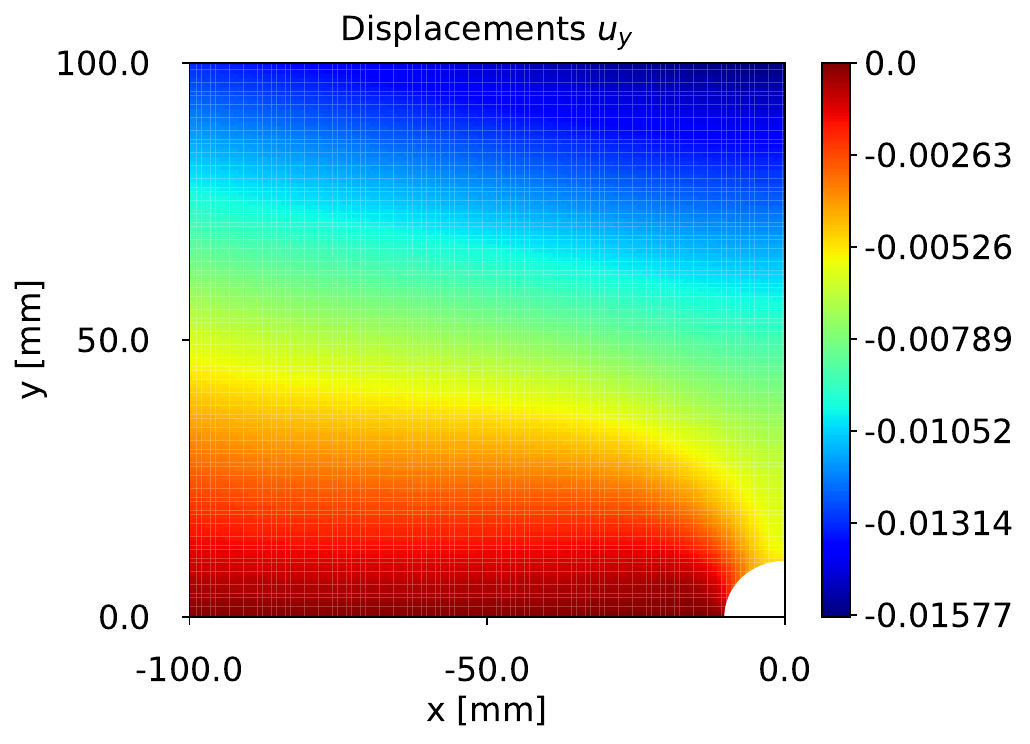}
    \caption{\ac{FEM} solution: Displacement field in $y$.}
    \label{fig:appendix_results_fem_uy.pdf}
  \end{subfigure}

  \begin{subfigure}[b]{0.49\linewidth}
    \includegraphics[width=\linewidth]{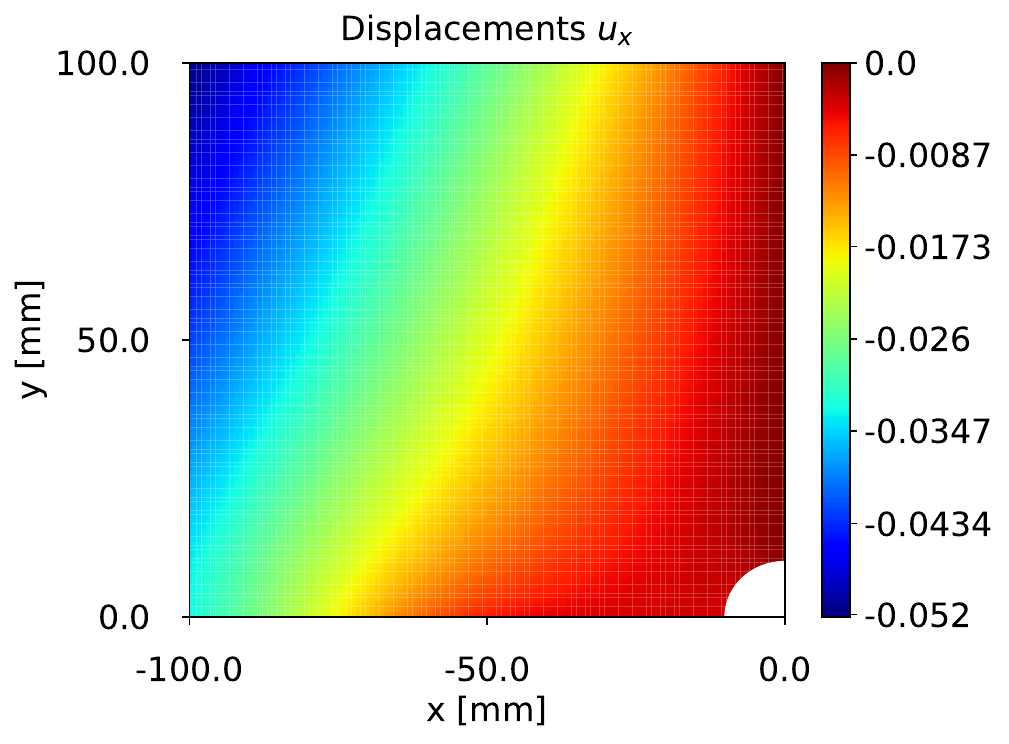}
    \caption{\ac{PINN} solution: Displacement field in $x$ without symmetry stress \acp{BC}, see \cref{fig:appendix_platewithhole_without_stress_symmetry_bcs}.}
    \label{fig:appendix_results_pinn_ux_without}
  \end{subfigure}
  \hfill
  \begin{subfigure}[b]{0.49\linewidth}
    \includegraphics[width=\linewidth]{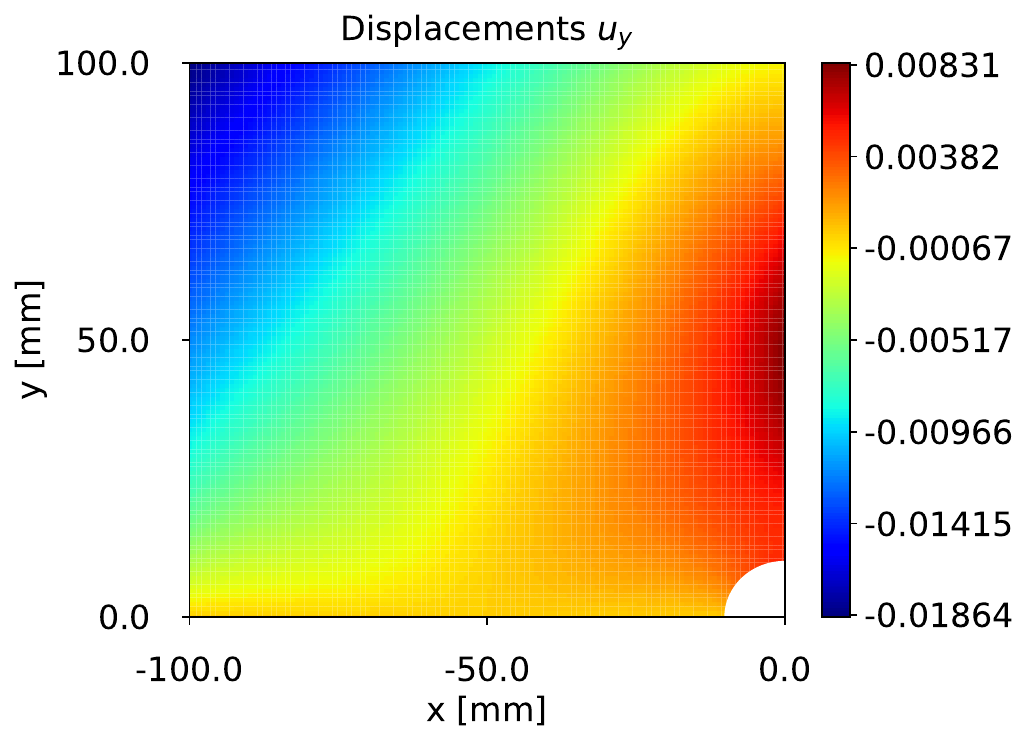}
    \caption{\ac{PINN} solution: Displacement field in $y$ without symmetry stress \acp{BC}, see \cref{fig:appendix_platewithhole_without_stress_symmetry_bcs}.}
    \label{fig:appendix_results_pinn_uy_without}
  \end{subfigure}

  \begin{subfigure}[b]{0.49\linewidth}
    \includegraphics[width=\linewidth]{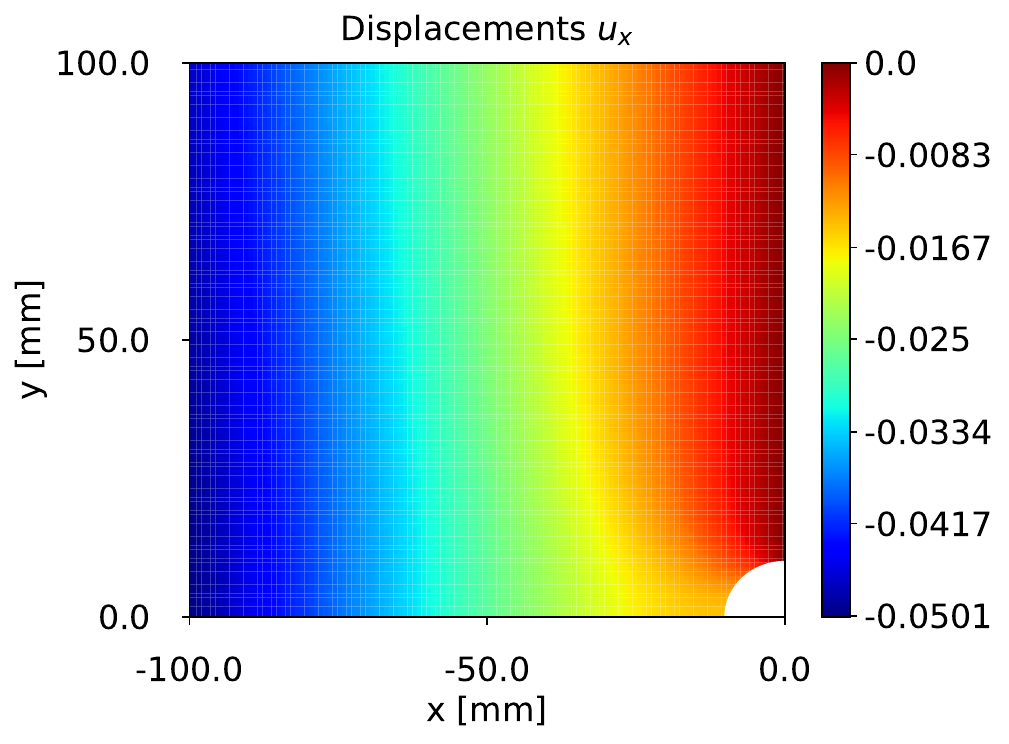}
    \caption{\ac{PINN} solution: Displacement field in $x$ with symmetry stress \acp{BC}, see \cref{fig:appendix_platewithhole_with_stress_symmetry_bcs}.}
    \label{fig:appendix_results_pinn_ux_with}
  \end{subfigure}
  \hfill
  \begin{subfigure}[b]{0.49\linewidth}
    \includegraphics[width=\linewidth]{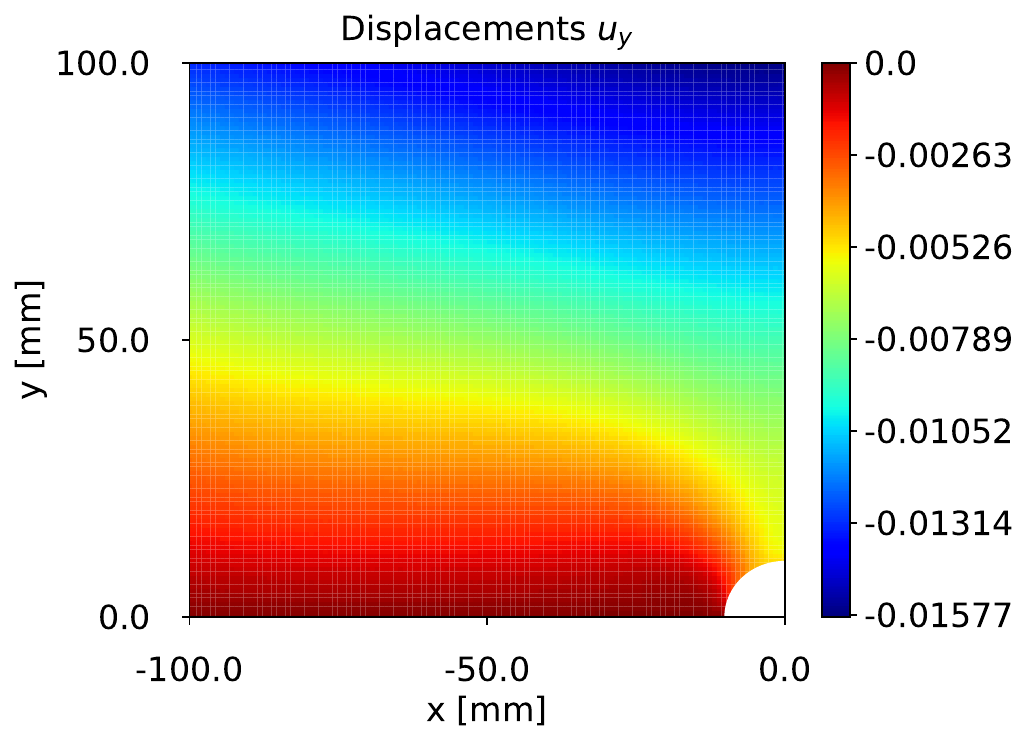}
    \caption{PINN solution: Displacement field in $y$ with symmetry stress \acp{BC}, see \cref{fig:appendix_platewithhole_with_stress_symmetry_bcs}.}
    \label{fig:appendix_results_pinn_uy_with}
  \end{subfigure}
  
  \caption{Resulting displacement fields for the test case plate with a hole with \acp{BC} as described in \cite{li_PINNsForElasticPlates_2021} (c, d) and our formulation presented earlier in \cite{henkes_PINNsForContinuumMechanics_2022} (e, f). The reference solution (a, b) is provided by a high-fidelity \ac{FE} simulation.}
  \label{fig:appendix_results_bc_comparison}
\end{figure}

The \acf{MAE} and the \acf{rL2} of the \ac{PINN} solution with and without symmetry stress \acp{BC} compared to a high-fidelity \ac{FE} solution are summarized in \cref{tab:appendix_bcs_errors}. For validation, we randomly select $\SI{2048}{\nounit}$ points from the \ac{FE} solution. In addition, we show the \ac{PINN} solutions we obtained with and without symmetry stress \acp{BC} as well as the \ac{FE} reference solution in \cref{fig:appendix_results_bc_comparison}.

\begin{table}[ht]
\centering
\caption{\Acf{MAE} and \acf{rL2} of the \ac{PINN} for the test case with and without symmetry stress \acp{BC} compared to a high-fidelity \ac{FE} solution.}
    \begin{tabular}{p{15mm}C{45mm}C{45mm}} 
    \toprule
                    & with symmetry stress \acp{BC}         & without symmetry stress \acp{BC}      \\ 
    \midrule
    \ac{MAE}        & \SI{5.3812 e-6}{\nounit}               & \SI{5.7706 e-3}{\nounit}             \\ 
    \midrule
    \ac{rL2}        & \SI{3.5649 e-4}{\nounit}               & \SI{3.7064 e-1}{\nounit}             \\ 
    \bottomrule
    \end{tabular}
\label{tab:appendix_bcs_errors}
\end{table}

\section{Error measures}\label{sec:appendix_error_measures}

In order to validate the performance of our parametric \ac{PINN} formulation, we compare the \ac{PINN} predictions to the solutions of high-fidelity \ac{FE} simulations. We consider the \ac{MAE} as an absolute error measure and the \ac{rL2} as a relative error measure. In the following, $\FEMSup{\vec{u}}\elm{2\numNodes}$ represents the vector containing the displacements of all $\numNodes$ nodes with coordinates $\{\vec{X}^{(i)}\}^{\numNodes}_{i=1}$ in the \ac{FE} discretization. The vector $\PINNSup{\vec{u}}\elm{2\numNodes}$ contains the displacements predicted by the parametric \ac{PINN} where the \ac{PINN} is evaluated according to \cref{eq:parameters_to_state_map} at the coordinates $\left\{\vec{X}^{(i)}\right\}^{\numNodes}_{i=1}$. The same material parameters $\bm{\kappa}$ are used for both the \ac{FE} simulation and the evaluation of the parametric \ac{PINN}. \\

\noindent The \textbf{\acf{MAE}} is then defined as
\begin{equation}\label{eq:mean_absolute_error}
    \MAE_{\vec{u}} = \frac{1}{2\numNodes} \sum_{i=0}^{2\numNodes} \left\vert \PINNSup{\sca{u}_{i}} - \FEMSup{\sca{u}_{i}}\right\vert,
\end{equation}
where $\left\vert \bullet \right\vert$ is the absolute value of the quantity $\bullet$. \\

\noindent The \textbf{\acf{rL2}} yields
\begin{equation}\label{eq:relative_l2_error}
    \rLTwo_{\vec{u}} = \frac{\norm{ \PINNSup{\vec{u}} - \FEMSup{\vec{u}} }}{\norm{ \FEMSup{\vec{u}} }},
\end{equation}
with $\norm{\bullet}$ denoting the $\LTwo$-norm.\\

For the statistical evaluation of the calibration results, we consider the \acf{ARE}. In addition to the mean, minimum and maximum \ac{ARE}, we also calculate the \acf{SEM} which gives us information about the scatter of the \ac{ARE}. Here, $\trueSup{\bm{\kappa}}$ represents the vector of true material parameters and $\identifiedSup{\bm{\kappa}}$ the vector of material parameters identified by using the parametric \ac{PINN} as parameters-to-state map in the deterministic \ac{NLS} calibration. \\

\noindent The \textbf{\acf{ARE}} for material parameter $\kappa_{i}$ is defined as
\begin{equation}\label{eq:absolute_relative_error}
    \ARE_{\kappa_{i}} = \frac{\left\vert \identifiedSup{\kappa_{i}} - \trueSup{\kappa_{i}} \right\vert}{\trueSup{\kappa_{i}}}.
\end{equation}

\noindent The \textbf{\acf{SEM}} with respect to a certain error measure, for instance, the \ac{ARE}, is then calculated as
\begin{equation}\label{eq:standard_error_of_mean}
    SEM_{\kappa_{i}} = \frac{\sigma_{\kappa_{i}}}{\sqrt{\numTests}},
\end{equation}
where $\numTests$ is the number of test cases on which the statistical evaluation is based and $\sigma_{\kappa_{i}}$ is the standard deviation, e.g., of the \ac{ARE}.


\clearpage

\bibliography{literature}

\end{document}